\documentclass[final,twocolumn]{elsarticle}

\usepackage{graphicx}
\usepackage{xcolor}
\usepackage{booktabs}
\usepackage{tabularx}
\usepackage{comment}
\usepackage{multirow}
\usepackage{colortbl} 
\usepackage{amsmath} 
\usepackage{color,soul}
\sethlcolor{yellow}

\usepackage{amssymb}

\begin{document}

\begin{frontmatter}



\title{Resilience of Vision Transformers for Domain Generalisation in the Presence of   Out-of-Distribution Noisy Images}


\author[inst1]{Hamza Riaz}

\affiliation[inst1]{organization={School of Computing, Dublin City University},
            addressline={Glasnevin Campus}, 
            city={Dublin},
            country={Ireland}\\
            alan.smeaton@dcu.ie}

\author[inst1]{Alan F. Smeaton}


\begin{abstract}

Modern AI models excel in controlled settings but often fail in real-world scenarios where data distributions shift unpredictably—a challenge known as domain generalisation (DG). This paper tackles this  limitation by rigorously evaluating vision transformers, specifically the BEIT architecture which is a model pre-trained with masked image modelling (MIM), against synthetic out-of-distribution (OOD) benchmarks designed to mimic real-world noise and occlusions. We introduce a novel framework to generate OOD test cases by strategically masking object regions in images using grid patterns (25\%, 50\%, 75\% occlusion) and leveraging cutting-edge zero-shot segmentation (via Segment Anything and Grounding DINO) to ensure precise object localisation.
Experiments across three benchmarks (PACS, Office-Home, DomainNet) demonstrate BEIT’s known robustness while maintaining 94\% accuracy on PACS and 87\% on Office-Home despite significant occlusions—outperforming CNNs and other vision transformers by margins of up to 37\%. Analysis of self-attention distances reveals that the BEIT’s dependence on global features, correlates with its resilience. Furthermore, our synthetic benchmarks expose critical failure modes: performance degrades sharply when occlusions disrupt object shapes (e.g., 68\% drop for external grid masking vs. 22\% for internal masking). 
This work provides two key advances: (1) a scalable method to generate OOD benchmarks using controllable noise, and (2) empirical evidence that MIM and self-attention mechanism, in vision transformers enhance DG by learning invariant features. These insights bridge the gap between lab-trained models and real-world deployment that offer a blueprint for building AI systems that generalise reliably under uncertainty.

\end{abstract}


\end{frontmatter}


\section{Introduction}
\label{sec:Introduction}

Computer vision models regularly fail to generalise when they are tried on out-of-distribution (OOD) data .
This means that they have reduced reliability as well as possible safety and fairness concerns thus limiting their deployment. It becomes apparent when we inject randomness and speculation during the prediction operation when trained models are tested on data from unseen domain distributions as often machine learning systems rely on training distributions which make them vulnerable to use. For example, real-world systems like autonomous vehicle navigation show a huge decline in performance when interacting with even partially different conditions and settings during model inference, compared to their training distributions. Different weather \cite{8917269}, day vs. nighttime lighting conditions \cite{8569387}, poses and positioning of objects \cite{8954212} are recurring reasons which affect such systems. 

In robotics when introducing visual distractions to agents, some methods are known to significantly decrease performance \cite{stone2021distracting}. A few possible reasons why ML models are  poor at lrstninh generalisation could be racial biases, texture statistics, or object backgrounds \cite{gulrajani2021in}. These factors reduces a model’s ability to capture causal factors related to potential variations in the data but instead fit the model to fake representations.  To tackle OOD and for the  development of real-world ML systems, these issues have been partially addressed in the literature.

As we know, domain generalisation can be controlled by three factors: dataset types, network architectures, and model selection criteria. To overcome OOD related challenges, much work has been done including  solutions like additional data collection for different domains or environments, adversarial learning, and data augmentation for the purpose of learning generalised invariances from the training domain \cite{Ganin2017, akuzawa2020adversarial}. Prior research has also proposed techniques to find hyperparameters which maximise performance on an external domain by measuring relatedness \cite{pmlr-v139-krueger21a}. In summary, invariant feature learning, data augmentation, meta learning, life long learning, and generative adversarial learning are related methods frequently found in the literature from which we can extract some highlights.

\begin{enumerate}
    \item An extensive study in \cite{gulrajani2021in} indicates that larger models have greater generalisation compared to smaller architectures.
    \item \cite{8953760, albuquerque2020improving, 9462394} show how self-supervised learning helps to improve domain generalisation of a model for OOD by learning generalised and transferable features.
    \item The removal of texture information also boosts domain generalisation \cite{wang2019learning}.
    \item Methods to ignore the texture of images and focus on shapes and ignore background noise could lead to better generalisation \cite{asadi2019towards}.
\end{enumerate}

\noindent 

The success of transformers in natural language processing has motivated many to use transformers for vision based tasks \cite{NEURIPS2021_652cf383,dosovitskiy2020image}. Reasons for that include that ViT contains more uniform representations in all the network layers, self attention creates global information during the early layers, and residual connections help to propagate features \cite{dosovitskiy2020image}. The lessons from the literature encourage us to explore the latest vision transformers for domain generalisation because vision transformers have most of these characteristics. 

In \cite{pmlr-v139-touvron21a} the authors introduced a teacher-student based learning strategy for efficient training on a single computer. Similarly, LeViT \cite{Graham_2021_ICCV} also focused on the computational advancements by using few CNN functions in transformers and creating hybrid architectures. BEIT~\cite{bao2022beit} is another state-of-the-art transformer-based method which has a denoising auto-encoder implementation to pre-train a vision transformer which is used in this paper. During pre-training, it randomly masks a number of patches with a proportion of the image and feed this corrupted input to the transformer which is one of the key points to tackle in OOD.

To address some of the issues related to OOD, this paper investigates aspects of DG using vision transformers. For proof of concept, we implemented a pipeline to check the OOD capability of four available pre-trained vision transformers. Originally each of these models were pre-trained and then fine-tuned on ImageNet-21k and ImageNet1k \cite{russakovsky2015imagenet} respectively. We then run inference on unseen benchmarks including the ImageNet-Sketch \cite{wang2019learning}, ImageNet-R (endition)~\cite{hendrycks2021many}, ImageNet Adversarial~\cite{hendrycks2021natural}, and ImageNet Corrupted~\cite{hendrycks2018benchmarking} benchmarks. This work discovers that BEIT outperforms all other vision transformers and to the best of our knowledge, properties like denoising, the self-attention mechanism, and self supervised fine-tuning could be the main reasons.

Based on initial results, we choose BEIT for further analysis where pre-trained weights are used as feature extractors along with attention masks. We then fine-tune three separate models on three popular benchmarks for domain generalisation namely PACS, Home-Office, and DomainNet. During the training process, our method implements the {\it training-domain validation set}, inspired by~\cite{gulrajani2021in} which means that 80\%\ of data in each domain will be used for training and validation, and the remaining 20\%\ from each domain will be combined and used for testing. 
We compute detailed analysis of the self-attention mechanism using attention distance metrics where we explore the latent space of models. Our method tries to explain where models which learn early global information then have better domain generalisation and where this is not effective. To measure generalisation, the results section considers metrics like accuracy, gap and precision, which show significant improvement. We also provide graphical visualisations of attention maps for OOD cases. 

In a similar manner, in this paper we  test a crucial claim of an already trained BEIT model which is that the model will also have denoising ability. Towards this aim, we design a paradigm in which we add noise to images namely a periodic masking grid overlaid on images in our test benchmarks. Additionally, the idea of grid masking is inspired from the original paper \cite {chen2020gridmask} and details about this approach are discussed in our methodology,  Section~\ref{sec:Methodology}. This leads to the generation of new datasets that models will have never seen previously and such datasets could serve as an OOD test set. Briefly, for each selected benchmark, our method creates four different variations of periodic grid markings based on the number of grids and the locations of these grids with respect to class objects in the images. Similarly, from these variations of generated image data which can be seen later in Figures~\ref{paper6_fig3}, \ref{paper6_fig4} and \ref{paper6_fig5}, each image will have three variatiations based on their mask and object occlusion ratio which will be set at 25\%, 50\%, and 75\%. Thus our method  creates 36 different kinds of benchmark to measure the behaviour of a model. 

From a broader view, our  mechanism designed to measure the resilience of DG has two main tasks. The first is to generate a new periodic grid masking around or outside the targeted objects or shapes within a test image where the most important information for correct classification of that image already exists. The second task consists of inference on these generated benchmarks in order to obtain summary results. 

It is important for this research to measure the ratio between the targeted shapes of given class objects within images and the grid masks so that we can investigate performance changes by blocking different proportions of the object area and we call this the occlusion ratio. To obtain the area of masks of objects present in the given images, our method uses a combination of pre-trained models like Segment Anything (SAM) \cite{kirillov2023segment} and Grounding DINO \cite{liu2023grounding} to extract the required masks with text-prompts in a zero-shot instance segmentation manner. Furthermore, we can use these masks to calculate the overlapping regions between the grid masks and objects, as  explained in sub-section~\ref{chap:generate_OOD_methodology}.

More importantly, by conducting these experiments we focus on the  measurement of variation or stretchiness we can induce into a data distribution and what type of effects we can expect in terms of domain generalisation. This paper also uses a model, generalised to new domains under all the conditions described earlier and not needing any transfer learning or fine-tuning to adapt it for new domains. The experimental results in Tables~\ref{paper6_tab1}, \ref{paper6_tab2}, and \ref{paper6_tab3}  illustrate this point.



To the best of our knowledge, our  method is a unique way to generate new benchmarks to test the OOD detection capabilities  by using zero shot instance masking. To design the basic framework, we implemented  simple grid masking functions to datasets and created grid masks of the same size as the original images and then used the popular SAM \cite{kirillov2023segment} and Grounding DINO \cite{liu2023grounding} to obtain precise masks of objects appearing in the images using  text-prompts. We then find overlapping regions between  grid masks and  object masks to determine  occlusion ratios. Details of the design method are given in Section~\ref{sec:Methodology}. 

\section{Related Work}

\label{sec:Related Work}
In the literature many methods have been proposed to address various challenges with OOD data,  including collection of extra data, invariant feature learning, data augmentation, meta learning, life long learning, and generative adversarial learning.  For instance \cite{Sagawa*2020Distributionally} is a  
distributionally robust group optimisation method which uses empirical risk minimisation and explains the importance of domains with error rates. Similarly \cite{li2018domain,sun2016deep} involves invariant feature learning across  domains which learns features which are invariant to external data variability. However, such methods were critcised in \cite{pmlr-v97-zhao19a} who highlighted reasons why invariant feature representation is insufficient for domain generalisation. 

Approaches based on invariant risk minimisation also exist where the network learns an optimal classifier on  top of invariant feature representations \cite{arjovsky2019invariant}. A popular meta learning based method known as Meta-Learning for Domain Generalisation (MLDG)  builds on model-agnostic meta learning where a meta learner is generalised on various domains.  To enhance generalisation, \cite{NEURIPS2018_1d94108e} explains that training on a single domain is sufficient when using adversarial data augmentation training.  

The above methods have their own strengths and weaknesses, but no single approach is best. Another concern is that they use a CNN as their backbone with pre-trained weights which means models will have a limited respective field compared to vision transformers.

Vision transformers are in principle more appropriate for generalisation than CNNs because of factors like global understanding, handling of variable-length inputs, fewer parameters, an attention mechanism, and pre-training.  The initial idea of the ViT \cite{dosovitskiy2020image} is straightforward. First, the image’s patches and positional embeddings after flattening go to a transformer encoder where multi-head attention helps the transformer with coverage with the respective loss function. In the same way, ViT and other vision transformers including LeViT, DeiT, BEIT \cite{dosovitskiy2020image,pmlr-v139-touvron21a,Graham_2021_ICCV, bao2022beit} followed similar architectures but pre-training requires much computational resources and large datasets. Similarly \cite{zheng2022prompt,Sultana_2022_ACCV} also investigated domain generalisation with vision transformers especially, ViT used as a backbone but in our work we  use BEIT.

Our approach leverages the Segment Anything Model (SAM) and Grounding DINO for generating OOD benchmarks using grid masking and offers distinct advantages over traditional generative models like GANs and diffusion models. GANs often face challenges such as mode collapse, where the generator produces limited and repetitive samples this leads towards a reduction in diversity in the generated data \cite{10.1007/978-981-97-2550-2_23}. Additionally, GANs can experience training instability and require extensive computational resources. While diffusion models address some of these issues by providing more stable training and high-fidelity samples, however, they still demand significant computational power and may not precisely control occlusion patterns. 

In contrast, our method utilises SAM and Grounding DINO to perform zero-shot instance masking with text prompts, enabling precise and controllable segmentation of objects without the need for extensive retraining. This approach maintains the original data distribution while systematically introducing occlusions. This creates more realistic and interpretable OOD benchmarks and by employing multiple grid variations, we can effectively assess a model's robustness to OOD scenarios which provides a scalable and efficient alternative to traditional generative techniques.

\paragraph{Segment Anything (SAM): } The paper ``Segment Anything" describes a versatile and powerful model that can execute segmentation tasks in a variety of contexts without requiring task-specific training. The authors describe a unique technique known as the Segment Anything Model (SAM), which uses a highly generalisable framework to effectively segment items in images \cite{kirillov2023segment}. SAM's primary notion is to accomplish high-performance segmentation through a mix of prompt engineering and a unique transformer-based architecture. The model operates by receiving prompts such as points, boxes, or masks as input and then creates the segmentation masks accordingly. SAM's architecture consists of an image encoder, a flexible prompt encoder, and a mask decoder, which processes an  input image and prompt to create the necessary segmentation output. The primary benefit of SAM is its capacity to adapt to multiple segmentation tasks without requiring explicit fine-tuning for each job, considerably increasing its usefulness.

\paragraph{SAM+Grounding DINO: } The original method for SAM only works with three types  of prompt modes namely bounding boxes, points and masks and is most effective and accurate with its bounding boxes prompt mode. By combining with another method like Grounding DINO \cite{liu2023grounding} which takes image and text prompts as input and outputs the bounding boxes according to the text prompts, these generated bounding boxes are given to the SAM model as a input prompts, along with the image and tasked to segment objects in the image. 

To describe Grounding DINO \cite{liu2023grounding}, the paper ``Grounding DINO: Marrying DINO with Grounded Pre-Training for Open-Set Object Detection" describes a novel approach that combines the DINO \cite{oquab2023dinov2} framework with grounded pre-training techniques to improve open-set object detection capabilities. Grounding DINO tries to overcome the limits of classic object identification models including SAM, which often perform well only on preset categories by allowing for the detection of objects outside of the training set. The essential innovation is the combination of DINO's strong implicit neural representations with grounded pre-training, which takes into account extra contextual information and different training data. This integration enables the model to generalise more effectively and to recognise a broader range of items in different circumstances. 

\paragraph{Masking Related Methods:}
Another component of the research in this paper is to conduct an investigation into available masking methods and  to select the most optimal  for our research. In the literature, Grid Masking \cite{chen2020gridmask}, Random Erasing \cite{zhong2020random}, Cutout \cite{devries2017improved}, Mixup \cite{zhang2017mixup}, CutMix \cite{yun2019cutmix}, Patch Masking \cite{he2022masked}, DropBlock \cite{ghiasi2018dropblock}, Image Inpainting \cite{pathak2016context}, Semantic Masking \cite{chen2017deeplab}, and Hide-and-Seek \cite{kumar2017hide}, are the most well known  which perform different types of masking. We now explain each of them briefly.

\begin{itemize}
    
\item \textbf{Grid masking} involves overlaying a grid on an image and randomly masking (obscuring) some of the grid cells \cite{chen2020gridmask}. 
\item The \textbf{Random erasing} method randomly selects a rectangular region in an image and replaces it with random values, zeros, or a constant value \cite{zhong2020random}. 
\item \textbf{Cutout} is similar to random erasing but specifically zeros out (or replaces with a fixed value) a randomly chosen square region in an image \cite{devries2017improved}. 
\item \textbf{Mixup} combines two images and their labels by taking a weighted sum of both. The pixels of the two images are mixed according to a mixing parameter, and the labels are mixed in the same proportion \cite{zhang2017mixup}. 
\item \textbf{CutMix} combines elements of Cutout and Mixup. It replaces a rectangular region of one image with a patch from another image while also mixing their labels accordingly \cite{yun2019cutmix}. 
\item \textbf{Patch masking} involves masking out random patches in an image. This method is particularly useful in self-supervised learning, where the model is trained to predict the masked patches, encouraging it to learn robust and meaningful representations of the data \cite{he2022masked}. 
\item \textbf{DropBlock} is a structured form of dropout where contiguous regions of the feature map are dropped during training \cite{ghiasi2018dropblock}. 
\item \textbf{Image inpainting} masks out a portion of an image and trains a model to reconstruct the missing parts \cite{pathak2016context}. 
\item \textbf{Semantic masking} involves masking specific objects or regions in an image based on their semantic meaning \cite{chen2017deeplab}. 
\item \textbf{Hide-and-Seek} is a data augmentation technique where parts of an image are randomly hidden during training \cite{kumar2017hide}. 
\end{itemize}

What all these masking methods have in common is that they have been developed to enhance the generalisation, diversity, robustness, and regularisation of any model during the time of training.

\section{Benchmarks Generations and Details }
\label{sec:Benchmarks}

This section highlights theoretical and visual details of benchmarks we have used in our experiments. There are three use cases for  benchmarks as this section explains,  benchmarks used in feasibility studies, benchmarks used in our analysis of vision transformers, and benchmarks used in the generalisation of new OOD benchmarks. 

\subsection{Benchmarks used in the feasibility study}
\begin{enumerate}
    \item The \textbf{ImageNet-Sketch}~\cite{wang2019learning} dataset has 50,000 images with 1K classes~\cite{ILSVRC15}, 50 images for each of the 1,000 ImageNet classes.
    \item \textbf{ImageNet-R (rendition)}~\cite{hendrycks2021many} contains 30,000 image renditions for 200 ImageNet~\cite{ILSVRC15} classes which is a subset of ImageNet-1K~\cite{ILSVRC15}. In addition, ImageNet-R (rendition) has distributions including art, cartoons, deviantart, graffiti, embroidery, graphics, origami, paintings, patterns, plastic objects, plush objects, sculptures, sketches, tattoos, toys, and video game renditions of ImageNet classes.
    \item \textbf{ImageNet-adversarial}~\cite{hendrycks2021natural} consists of adversarially filtered real-world images to fool ImageNet classifiers and it also contains 200 classes as a subset of ImageNet-1K.
    \item \textbf{ImageNet Corrupted}~\cite{hendrycks2018benchmarking} consists of images with 75 common visual distractions and 1,000 classes and the goal is to improve and evaluate the robustness of models. 
    ImageNet-Corrupted (ImageNet-C) is a dataset created to evaluate the robustness of image classification models against common corruptions and perturbations. The dataset includes 15 different types of corruption, each applied at 5 severity levels. These corruptions can be grouped into four categories: \textbf{Noise:} Gaussian noise, shot noise, impulse noise. \textbf{Blur:} Defocus blur, frosted glass blur, motion blur, zoom blur. \textbf{Weather:} Snow, frost, fog, brightness. \textbf{Digital:} Contrast, elastic transformation, pixelation, JPEG compression
\end{enumerate}

\begin{figure*}[htbp]
\centering
\includegraphics[width=0.6\textwidth]{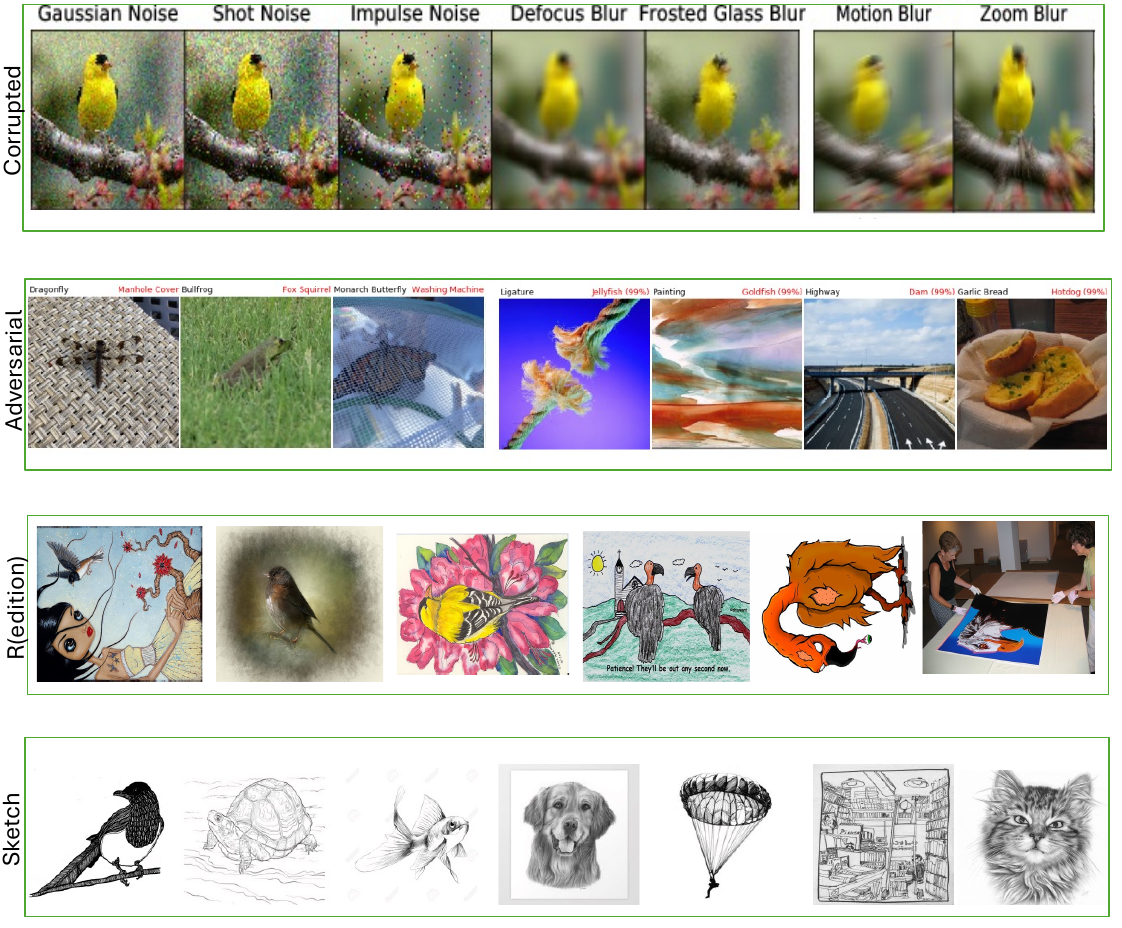} %
\caption{Visual illustration of selected OOD benchmarks, different versions of ImageNet, to test the robustness of models.}
\label{paper5_fig1}
\end{figure*}

\noindent 
Fora clearer indication of  the variations in these benchmarks, a visual summary is presented in Figure~\ref{paper5_fig1}. Each row indicates the type of dataset and we can see that each data distribution is quite different with respect to others which explains the variety for measuring domain generalisation. Selected vision transformer models have never seen these distributions during their pre-training and fine-tuning procedures. 

\subsection{Original benchmarks  used for exploration}

\begin{table*}[ht!]
\caption{Benchmarks used  for the analysis of vision transformers for domain generalisation} 
\centering 
\small
\begin{tabular}{lcccp{3cm}c} 
\toprule 
\textbf{Datasets} & \textbf{Domains} & \textbf{Classes} & \textbf{Samples} & \textbf{Descriptions} & \textbf{References} \\ [0.5ex] 
\midrule 
Office-Caltech & 4 & 10 & 2,533 & Caltech, Amazon, Webcam, DSLR & \cite{10.1007/978-3-642-15561-1_16} \\ 
Office-31 & 3 & 32 & 4,110 & Amazon, Webcam, DSLR & \cite{10.1007/978-3-642-15561-1_16} \\
PACS & 4 & 7 & 9,991 & Art, Cartoon, Photos, Sketches & \cite{li2017deeper} \\
VLCS & 4 & 5 & 10,729 & Caltech101, LabelMe, SUN09, VOC2007 & \cite{6751316}\\
Office-Home & 4 & 65 & 15,588 & Art, Clipart, Product, Real World & \cite{venkateswara2017Deep}\\ 
Terra Incognita & 4 & 10 & 24,788 & Wild animal images recoded at four different locations L100, L38, L43, L46 & \cite{Beery_2018_ECCV}\\
Rotated MNIST & 6 & 10 & 70,000 & Rotated hand-written digits & \cite{ghifary2015domain}\\
DomainNet & 6 & 345 & 586,575 & Clipart, Infographs, Paintings, Quickdraw, Real, Sketch & \cite{Peng_2019_ICCV} \\[1ex] 
\bottomrule 
\end{tabular}

\label{table:nonlin} 
\end{table*}
Table~\ref{table:nonlin} presents a summary of some of the open DG benchmarks used in the literature for supervised learning. For our initial experiments, we explored two of these, PACS and Office-Home. 
In the case of reinforcement learning, RoboSuite, DMC-Remastered, DMC-GB, DCS, KitchenShift, NaturalEnvs MuJoCo, CausalWorld, RLBench, Meta-world and others are also commonly used benchmarks \cite{kirk2023survey}.

We restrict our work to vision-based benchmarks to narrow the scope because if we were to work with benchmark datasets across multiple domains covering vision, robotics, language processing, etc. then our results would have an overhanging question of whether results would have had as much to do with the domains chosen. 

Of the two benchmarks we use, one is a relatively simple dataset (PACS) with 4 different domains with images drawn from  Art, Cartoon, Photos, and Sketches. Each domain has 7 classes and there are 9,991 sample images in total. The second benchmark is Office-Home, also consists of images. This also has 4 domains namely Art, Clipart, Product, and Real World with 65 classes in each domain. The idea behind choosing the first is to work on a benchmark which could have comparatively less complex classification tasks so that we can observe the behaviours of domain-specific and domain-generic models. We select Office-Home as a second benchmark because of the higher number of classes (supervised tasks) which adds complexity to the tasks for each domain. 

\subsection{Generation of New OOD Benchmarks} \label{data_gen}

Here we explain  variations of newly generated OOD benchmarks which we created in order to carry out further experiments. For example, Figures~\ref{paper6_fig3}, \ref{paper6_fig4} and \ref{paper6_fig5} each show  two types of variation. On the y-axis, these figures have variations according to the number of grids and the locations or positions of those grids while the x-axis shows changes according to occlusion ratios which are 25\%, 50\%, and 75\%. 

For a 25\% occlusion ratio between object masks and grid masks, we considered the original grid pattern as shown in Figure~\ref{paper6_fig1}. This means that we  keep one grid per unit or window of grids. For a 50\% occlusion ratio, instead of getting simple grids, we use a checkerboard pattern. This means that we activate opposite but connecting grids yielding 50\% blocking of objects. Similar to the 50\%, the 75\% occlusion ratio also has a checkerboard pattern for grids but it allows overlapping between the grids of each unit or window.

Another important variation lies on the end of the y-axis of thee Figures which is related to generating grids outside the shapes of objects. This will distract models as with the occlusion ratio, the shapes of objects or silhouettes are  also deformations with this given style of grids. To calculate grids outside the shapes, after computing the masks of objects, we simply locate the bounding box coordinates according to masks and then generate the grids styles within that rectanglular area. Next we  highlight how these data generation approaches  look like for each of the three benchmarks.

\subsubsection{PACS Occlusion Benchmarks}
\begin{figure}[htbp]
\centering
\includegraphics[width=0.5\textwidth]{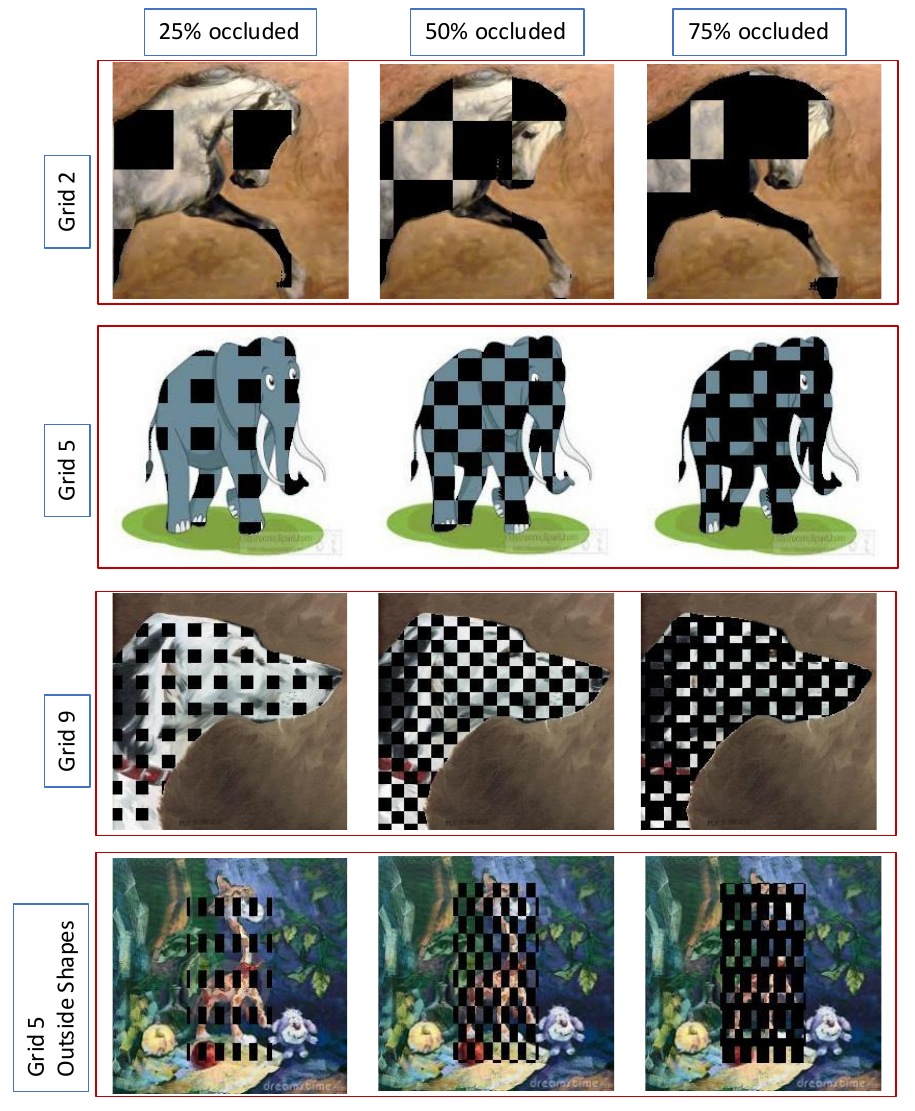} %
\caption{Generated data distributions for PACS with 12  variations to measure the resilience of the BEIT model. The y-axis shows types of newly generated distributions based on the number of grids and the x-axis shows different occlusion ratios.}
\label{paper6_fig3}
\end{figure}

This part of the paper shows OOD data generation for the PACS benchmark. We run our method on the testing set with only 1,014 images, the idea being to generate unseen data distributions which means this data is unseen by the model and not used in training. Therefore  adding these distractions to a dataset can also produce new unseen distributions. Figure~\ref{paper6_fig3} describes the visual results for each  variation on some of the different domains of the PACS dataset. 
For instance, in the case of a 2-grid system, we will have larger sized grids  to block the information and by increasing the number of grids, the size of grid masks decreases. 

\subsubsection{Office-Home Occlusion Benchmarks}

\begin{figure}[htbp]
\centering
\includegraphics[width=0.5\textwidth]{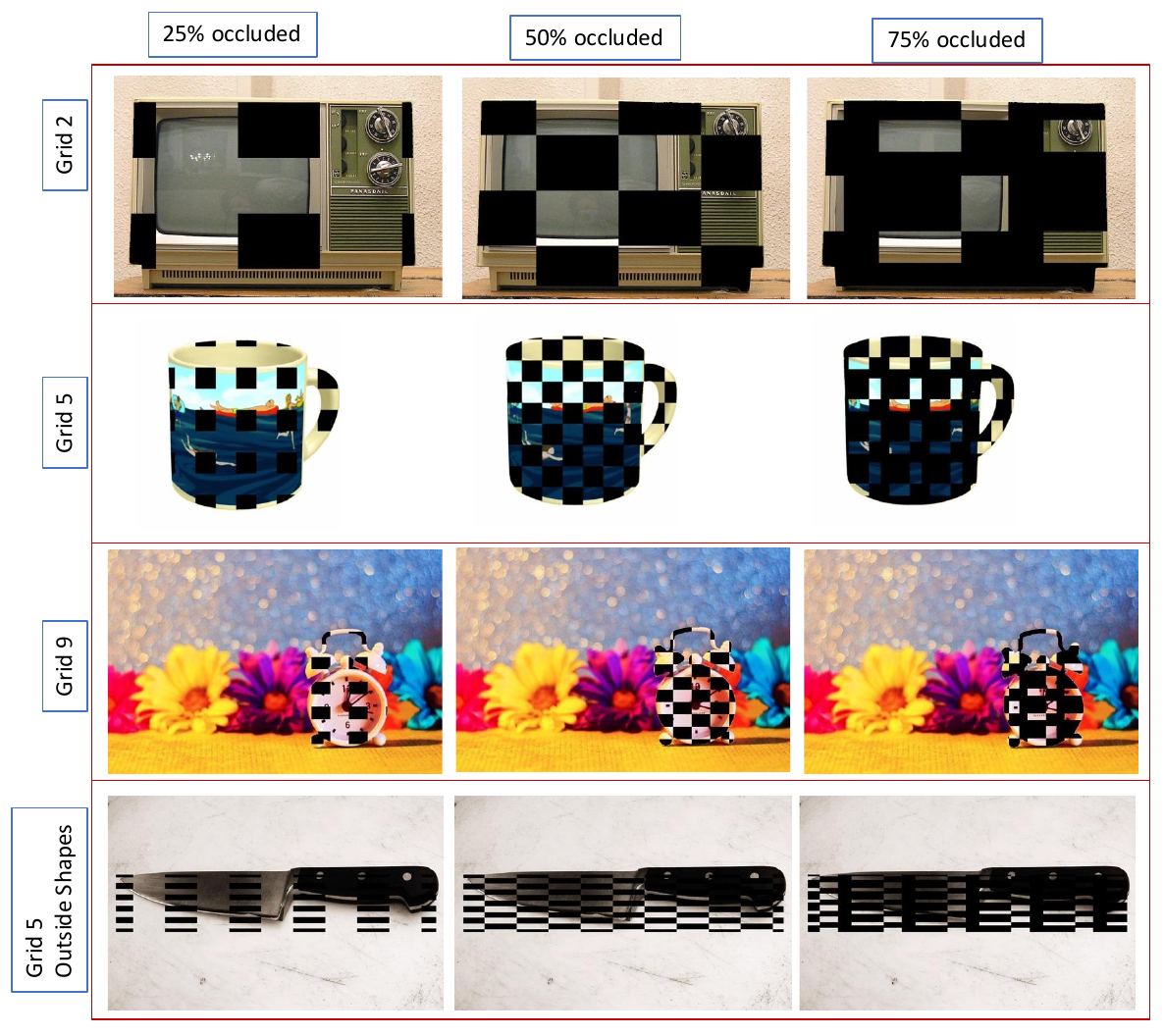} %
\caption{Generated data distributions for Office-Home with 12  variations to measure the resilience of the BEIT model. The 25\% means simple grids, 50\% means a checkerboard pattern, and 75\% means checkerboard with overlapping between the units.}
\label{paper6_fig4}
\end{figure}

When our  method is implemented on the second benchmark which is Office-Home, it has a much greater number of classes to work on. The results of some sample variations are shown in Figure~\ref{paper6_fig4}. Our data generation method for Office-Home follows the same settings as in PACS  data generation and for different domains we highlight the visual definitions in Figure~\ref{paper6_fig4}. Details of the original dataset are described in Table~\ref{table:nonlin}. The size of each unit depends on the number of grids and overall height and width of the input image. Our  method is implemented on the test-set given by the original dataset paper and the number of images this is applied to is 3,117. 

\subsubsection{DomainNet Occlusion Benchmarks}
\begin{figure}[htbp]
\centering
\includegraphics[width=0.45\textwidth]{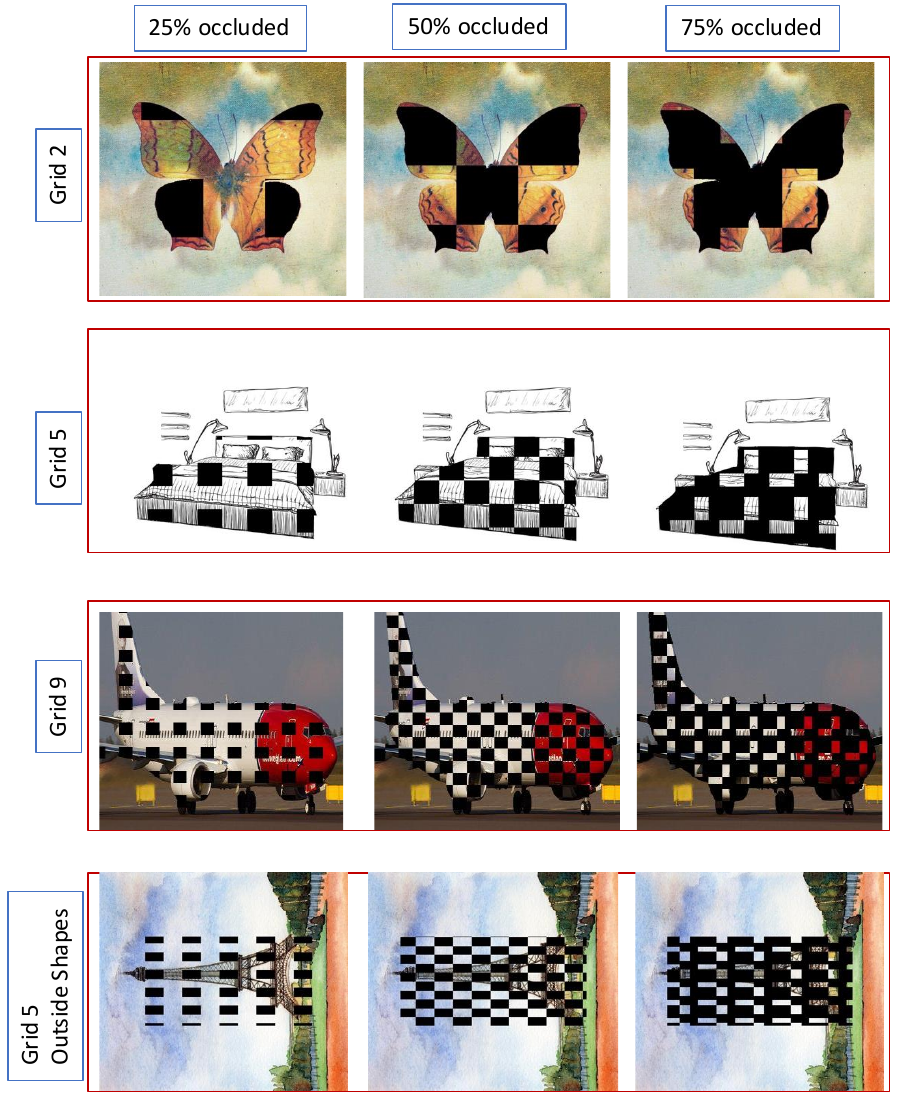} %
\caption{Generated data distributions for DomainNet with 12  variations to measure the resilience of the BEIT model.}
\label{paper6_fig5}
\end{figure}

We know that DomainNet is already a large-scale domain generalisation dataset with 6 different domains and more than 0.5 million images. The method we designed for creating additional testing images was implemented on 119,202 images and generated 12 more versions of the DomainNet dataset. Figure~\ref{paper6_fig5} illustrates how each variation blocks or masks the image information. 

\section{Methodology}
\label{sec:Methodology}
We now introduce the  model development and  experiments we conducted in our research. 

\subsection{Feasibility study}
The results of a feasibility study that we  conducted as  preliminary work \cite{riaz2023domain} are shown in Table~\ref{tab1} confirming the stable performance of various transformer-based models on OOD benchmarks.
Table \ref{tab1}  shows  OOD scores in the form of top-1 and top-5 accuracy for four  benchmarks namely ImageNet-Sketch \cite{wang2019learning}, ImageNet-R (endition)~\cite{hendrycks2021many}, ImageNet Adversarial~\cite{hendrycks2021natural}, and ImageNet Corrupted~\cite{hendrycks2018benchmarking}. Transformers including ViT~\cite{dosovitskiy2020image}, LeVit~\cite{Graham_2021_ICCV}, DeiT~\cite{touvron2021training}, and BEIT~\cite{bao2021beit} are tested and we took the baseline models from the huggingface library. All these models are pre-trained using ImageNet 21K classes~\cite{ILSVRC15} and fine-tuned with ImageNet2012 1k datasets ~\cite{ILSVRC15}. These models have never seen such diverse domains during their training processes yet are still able to classify with good confidence scores. Among all the models we observe that BEIT  outperformed all others which encourages us to correlate the unique points of BEIT with factors such as its mask image modelling with self-supervised learning of large models, a self-attention mechanism, and denoising of corrupted inputs.



\begin{table*}[ht]
\caption{Results using 4 vision transformers (rows) on 4 OOD related benchmarks (columns)} 
\centering 
\resizebox{1.7\columnwidth}{!}{%
\begin{tabular}{lcccccccc}\toprule& \multicolumn{2}{c}{ImageNet-Sketch} & \multicolumn{2}{c}{ImageNet-R(edition)}& \multicolumn{2}{c}{Imagenet Adversarial} & \multicolumn{2}{c}{Imagenet Corrupted}
\\
\midrule
Models           & Top1 Acc   &  Top5 Acc & Top1 Acc   &  Top5 Acc& Top1 Acc   &  Top5 Acc& Top1 Acc   &  Top5 Acc \\
\midrule
ViT    &35.43& 57.29 &32.82& 47.54 &12.97& 30.04& 78.06& 94.43   \\
LeViT  & 0.95 & 0.72 &0.81 & 0.44 & 9.13 & 27.14 & 73.67 & 90.98   \\
DeiT & 32.58 & 50.21 & 31.04 & 44.42 & 9.97 & 24.31 & 77.95 & 92.56  \\
\color{blue}BEIT   &\color{blue}47.55 & \color{blue}71.01 & \color{blue}44.72 & \color{blue}62.13 &\color{blue} 22.60 &\color{blue} 47.74 & \color{blue}81.88 &\color{blue} 96.41  \\
\bottomrule
\label{tab1} 
\end{tabular}
}
\end{table*}


\subsection{Fine-tuning of the BEIT Model}
Based on the numbers of domains, images, and classes, we chose to use PACS~\cite{li2017deeper}, Office-Home~\cite{venkateswara2017Deep}, and DomainNet~\cite{Peng_2019_ICCV} to measure the domain generalisation capability of BEIT for small, medium and large scale datasets. Further details of benchmarks can be found in~\cite{10.1007/978-3-031-37963-5_60}. During the training and validation steps, pre-processing includes image resizing, random horizontal flip, and normalisation. Similarly, testing includes re-sizing, centre cropping, and normalisation as pre-processing procedures. 

Inspired by the original paper~\cite{bao2022beit}, we used the based version of the BEIT transformer in this research. This has 12 transformer layers with 768 hidden and 3,072 feed-forward networks. Each attention layer has 12 attention heads of size 64 and these are responsible for learning self-attention masks.  Each image was divided into $14 \times 14$ patches of $ 16 \times 16$ pixels. BEIT is trained with 8,192 visual tokens.

\subsubsection{Feature Extraction and Fine-Tuning of Model}

We used the base BEIT model from the hugging face repository as a feature extractor. To extract features from input images, the model re-sizes images to $224 \times 224 \times 3$ resolution and the output features also have the same size. The model first converts images into $14 \times 14$ patches with a resolution of $16 \times 16$ for each and then flattens the patches and feeds them into the transformer with positional embeddings. The model is pre-trained and fine-tuned on ImageNet21k. After extracting features our method normalises feature maps with respect to the mean and standard deviation of images. We also extract an attention mask to ensure the self-attention mechanism of self-supervised BEIT is able to separate semantic regions and object boundaries. 

After normalising features for each benchmark, we configured the BEIT transformer model \cite{bao2022beit} to fine-tune it. It is clear that pre-training of a vision transformer is expensive so we froze most of the model and retrained only the last layers where we used Adam as the optimiser function with learning rate of 0.00005, weight decay of 0.05 and cross-entropy as the loss function. These benchmarks have vital differences in classes, complexity and samples~\cite{10.1007/978-3-031-37963-5_60}, hence for a fair analysis we trained three separate models for each benchmark. To avoid over-fitting, we added early stopping to training by monitoring losses with patience 5. Our method saves the best weights during each checkpoint. Each model has 85.8 million trainable parameters, and training and inference were executed using one RTX 3090 GPU and all experiments were performed with batch size of 8.

During the model development process, we adopted the {\it training-domain validation set} method to divide datasets into training, validation, and testing sets. Our method used 80\%\ of the data from each domain for training and validation and the remaining 20\%\ from each domain is combined as an overall testing set. Ideally, to test the OOD generalisation of a model, the testing set should be from other unseen domains but this overall test set is also unique and unseen to the BEIT transformer model. 

 The version of pre-trained weights which we used were pre-trained and fine-tuned on ImageNet 21k. For fine-tuning on a downstream task like classification, we appended task layers in the BEIT model and fine-tuned  parameters on the specific benchmarks for OOD domain generalisation. Like the original work, our method also uses average pooling to aggregate information, and then feed it to a softmax-based classifier. As an overview, our approach first does image pre-processing and extracts feature maps along with attention masks from frozen BEIT transformer layers for PACS~\cite{li2017deeper}, Office-Home~\cite{venkateswara2017Deep}, and DomainNet~\cite{Peng_2019_ICCV}. To perform a downstream task of classification we further fine-tuned the last layers of the model in addition to the pooling and classification layer for domain generalisation.  Figures~\ref{paper5_fig2} provide an overview of our results in term of accuracies and losses.
 
\subsubsection{Inference on ODD Benchmarks}
Following fine-tuning of hyperparameters for OOD datasets, our approach runs inference on a well-trained and shallow network on the unseen testing set from each benchmark. PACS has 9,991 images with 4 domains and 7 classes which is a comparatively  smaller dataset. Office-Home has 15,588 images also with 4 domains but it has 65 classes.  DomainNet is a larger benchmark with more than 0.5 million images, 6 domains and 365 classes. Although fine-tuning of any vision transformer is a  less time-consuming process than pre-training from scratch, this also depends on the size of the dataset. For instance, PACS and Office-Home take almost 4-6 hours for fine-tuning but in the case of DomainNet our model takes almost 3 days on the same equipment.

\begin{figure*}[htbp]
\centering
\includegraphics[width=0.9\textwidth]{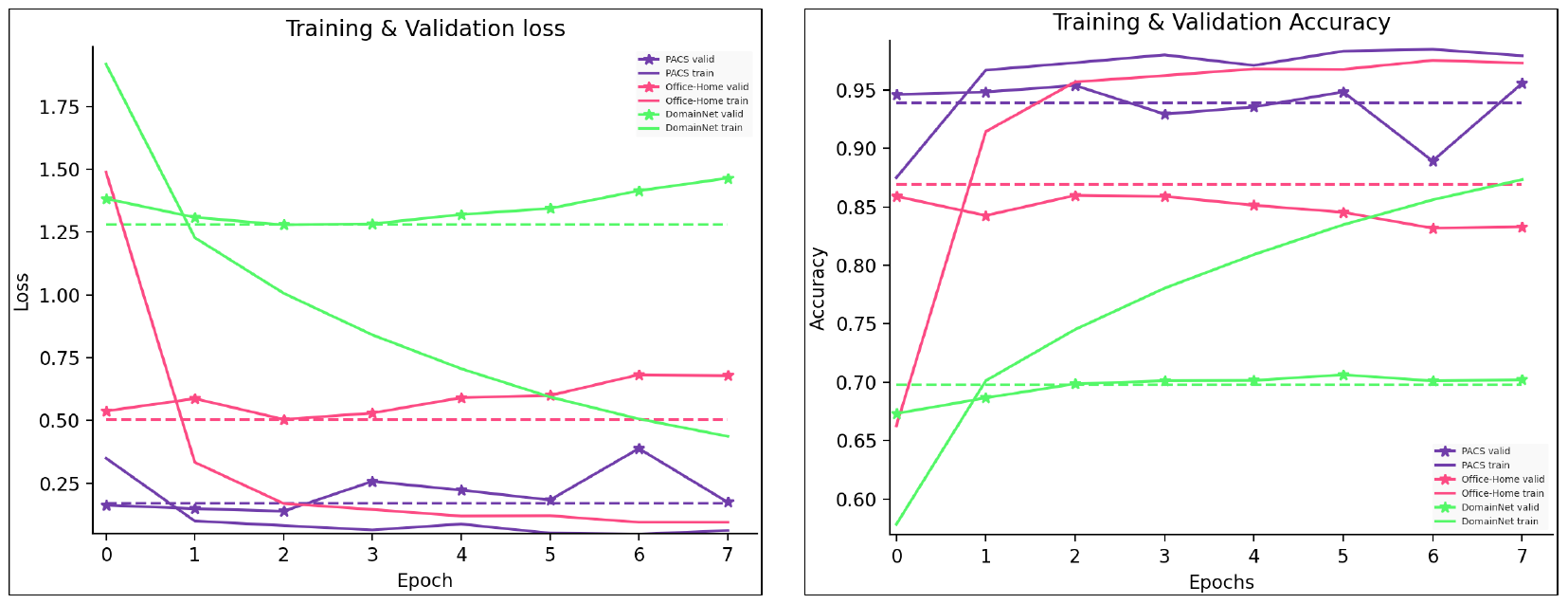}
\caption{Representation of accuracy and loss}
\label{paper5_fig2}
\end{figure*}

During the development of our model, we monitored accuracy and loss metrics for all three benchmarks and Figures~\ref{paper5_fig2}show our model fine-tuned for only 7 epochs because of our conditioning like early stopping and checkpoints for saving the best weights. The three different colours in the Figures represent the performance of each model and different line styles describe loss and accuracy for train, validation, and testing sets. Another important factor to consider in the graphs is the gap between the target and validation curves and this gap could directly  describe the domain generalisation ability of a model. For example, PACS shows a stable trend because the gap between target and validation curves is small for both  metrics which means the model has better domain generalisation for PACS. On the other hand, DomainNet has larger differences between train and validation curves but the gap between target and validation is small which means that in this case the BEIT vision transformer needs more investigation. The numerical results in Table~\ref{tab3} also reflect the same information.


\subsection{Attention Mechanism and Calculation of Attention Distances}

In self-attention layers, the mechanism determines the significance of each location (patch) in relation to a query patch by assigning attention weights. These weights represent how much influence each location has on the query patch. To analyse the behaviour of self-attention, the spatial distances between the query patch and the attended patches is calculated, with each distance weighted by the respective attention scores. By averaging these weighted distances across multiple datapoints, the average attention distance is obtained for each attention head. This metric provides insight into whether the attention head primarily aggregates local information (short distances) or global information (long distances), helping to distinguish patterns in information aggregation across the network.
The given pixel distances of each attention head are calculated by weighting them with the associated attention weights. These weighted lengths are then averaged across N data points to calculate the average attention distance for each head. The study demonstrates that lower levels show a combination of local (short distances) and global (long distances) attention, whereas upper layers mostly focus on global attention across the spatial domain.
To calculate attention distances in BEIT, for a single attention head $h$, the mean attention distance can be calculated by the following formula:
\begin{equation}
d_h = \frac{1}{N} \sum_{i=1}^{N} \sum_{K} W_h(Q_i, K) \cdot D(Q_i, K)
\end{equation}

\noindent 
Where $Q$ represents the query patch position, $K$ is the key of patch position in the spatial grid, $W_h(Q_i, K)$ is the attention weight given to $K$ for a query $Q_i$. $D(Q_i, K)$ is the Euclidean distance between $K$ and $Q_i$. The inner summation aggregates the weighted distances for a query $Q_i$ across all attended locations $K$. The outer summation averages the attention distances over all $N$ datapoints.

\subsection{Framework to Generate Object-Occluded Benchmark} \label{chap:generate_OOD_methodology}

\subsubsection{Implementation of Grid Masking}
Grid masking is a data augmentation approach in computer vision that improves neural network stability and generalisation. GridMask is a novel information removal method that employs structured dropping of uniformly distributed square regions, which contrasts with previous methods like Cutout and Hide-and-Seek. Unlike Cutout, which removes large continuous regions, or Hide-and-Seek, which randomly selects squares, GridMask deletes spatially uniform squares, allowing for better statistical balance between preserving and removing information. The GridMask which we implemented in our  method is inspired by the original work in~\cite{chen2020gridmask} and has the following components:

\begin{itemize}
    \item \textbf{Grid Overlay:}
    The image is covered with a grid of fixed-size cells. Each cell represents a tiny rectangle or square portion of the picture.
    \item \textbf{Random Masking:}
    A fraction of these grid cells are randomly chosen to be masked. The selection might be based on a preset probability or a predetermined number of maskable cells.
    \item \textbf{Masking Operation:}
    The selected grid cells are then hidden or ``masked." This may be done in one of a number of different ways including \textbf{Zero Masking:} Set the pixel values in the chosen cells to zero.
\textbf{Random Values:} Replace the pixel values with random values that are evenly distributed or derived from a normal distribution.
\textbf{Constant Value:} Replace the pixel values with a constant value (such as the dataset's mean pixel value).
    \item \textbf{Augmented Image:}
    The end result of this is an enhanced image with specific sections defined by grid cells blocked off. This image is then utilised as input while training the neural network.
\end{itemize}

\begin{figure}[htbp]
\centering
\includegraphics[width=0.35\textwidth]{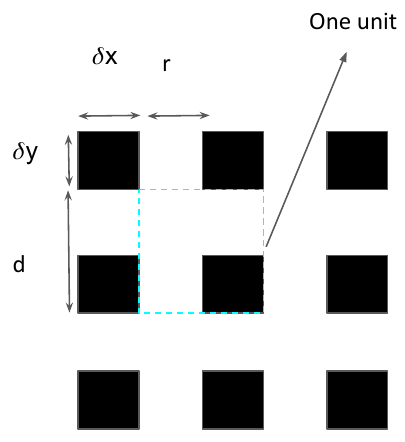} %
\caption{The basic diagram for GridMask and how one unit of the grid appears.
\label{paper6_fig1}}
\end{figure}

\noindent 
Mathematically the setting of GridMask method can be written as, 
\[
\tilde{x} = x \times M
\]
where $x \in \mathbb{R}^{H \times W \times C}$ represents the input image, $M \in \{0, 1\}^{H \times W}$ is the binary mask that stores pixels to be removed, and $\tilde{x} \in \mathbb{R}^{H \times W \times C}$ is the result produced by the algorithm. For the binary mask $M$, if $M_{i,j} = 1$ it keeps pixel $(i, j)$ in the input image otherwise it removes it. Figure~\ref{paper6_fig1} illustrates the fundamental concepts behind the GridMask method, which we used in our approach.

Four numbers are used by this method. $r$ is the ratio of the shorter white edge in a unit. $d$ is the length of one unit. $\delta x$ and $\delta y$ are the distances between the first intact unit and boundary of the image.

\[
k = \frac{\sum (M)}{H \times W}
\]
As $r$ computes the keep ratio of the input image therefore we first calculate $k$ for a given mask $M$ and then $r$ can be calculated using the following equation. 
\[
k = 1 - (1 - r)^2 = 2r - r^2
\]
After finding the keep ratio, the algorithm determines $l$. Therefore, the length of one unit $d$ does not impact the keep ratio. But it determines the size of a single dropped square.
When $r$ is fixed, the relationship between one dropped square's side length $(l)$ and $d$ is
\[ l = r \times d \]
The larger $d$ produces larger $l$. Hence to add randomness, $d$ is selected randomly using the given equation.

\[ d = \text{random}(d_{\text{min}}, d_{\text{max}}) \]

\noindent 
In the end, $\delta x$ and $\delta y$ can shift the mask according to given $r$, and $d$ and covers all  possible situations. $\delta x$ and $\delta y$ are also selected randomly.

\[
\delta x(\delta y) = \text{random}(0, d - 1)
\]

\subsubsection{Working with SAM and Grounding DINO for Object Masking}
\begin{figure*}[!htb]
\centering
\includegraphics[width=0.85\textwidth, angle=180]{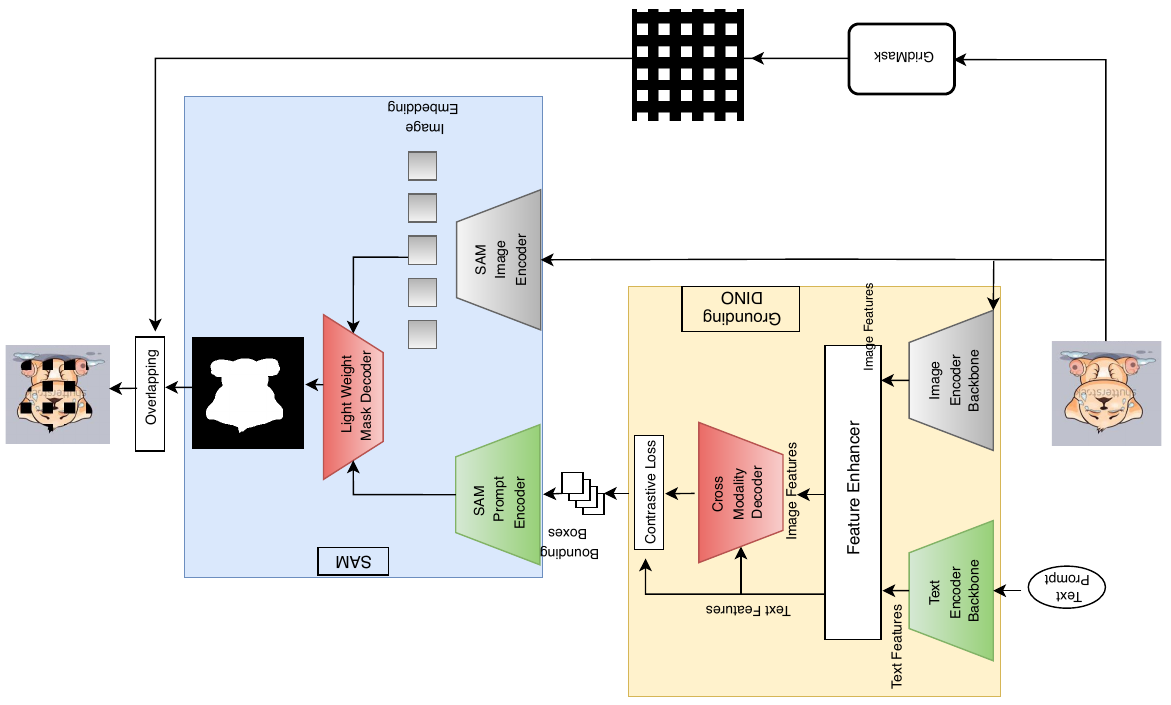} %
\caption{Overview diagram illustrating how SAM and DINO combine to segment objects in images in our test sets.
\label{paper6_fig2}}
\end{figure*}

Our goal is to generate an object-occlusion-based grid masking which is itself a unique concept to block regions of interest with the most relevant information. To fulfil this goal, the first step will be to calculate the grid marking for whole input image as shown in Figure~\ref{paper6_fig2} by using fundamentals from Section~\ref{chap:generate_OOD_methodology}. 

In parallel to the above, our  method with cascade settings also takes an input image with a corresponding text label as a prompt. In this cascade of two models, we first  use the GroundingDINO model which will boost the power of the original SAM model. The main motivation for using GroundingDINO is to add a more interactive mode into SAM, namely the use of text prompts. In the original paper describing SAM, although the text prompt mode is mentioned, it is not implemented in the official code base for the technique. Therefore, we designed our own mechanism to get the best of both models in such cascade settings. 

At the start of this process, each of the input images in the benchmark dataset is given to a backbone image encoder and related text prompts are given to the text encoder backbone, as shown earlier in Figure~\ref{paper6_fig2}. Each of the two encoders extract basic image and text features respectively and we then apply another important function ` known as a feature enhancer. In addition, the cross-modal decoder in GroundingDINO provides the first bridge to correlate  textual information to image features. The contrastive loss function takes input from the cross-modality decoder and the new text features and  yields localisation output in the form of bounding boxes. 
In Figure~\ref{paper6_fig2}, the processes shown in  yellow  relates to GroundingDINO.

As we have already established that SAM works better in bounding box mode which means that bounding boxes will be given to SAM as input prompts, we use the first model to get the bounding boxes of a given class and the input images will again pass into the SAM image encoder. This will generate image embeddings and in the end a lightweight decoder will accept the embeddings of bounding boxes and embeddings of images to produce segmentation masks which can be seen in Figure~\ref{paper6_fig2}. This mask will be the final output of the cascade network with a zero shot instance segmentation. Therefore, by combining both methods in a cascade manner it can enable generalised zero shot settings and we can deliver a segmentation model based on zero shot text-prompts.

\subsubsection{Finding Overlapping Regions of Interest}
After determining the object masks in all the images in each of our benchmark datasets, the next step in the process is  to  compute overlapping regions between grid masks and object masks. In the method we developed, to calculate and keep only the grids from a grid mask that overlap with an object mask, we need to perform a number of  steps. For example,
let \( G \) be the grid mask and \( O \) be the object mask, where \( G, O \in \{0, 1\}^{H \times W} \), which means both will be binary masks.
We can calculate the overlapping region mask:
\[G_O = G \odot O
\]
where \( \odot \) denotes the element-wise multiplication, keeping only the grids present in the overlapping regions \( G_O \). To get our final results, this new grid mask \( G_O \) will be further multiplied by the original image and the end result is shown in Figure~\ref{paper6_fig2}.

Let \( I \) be the original image where \( I \in \mathbb{R}^{H \times W \times C} \), to extend the mask across channels:

\[G_O^{(ext)} = G_O \otimes \mathbf{1}_C
\]
where \( \otimes \) denotes the outer product with a vector of 1's of length \( C \). Then, apply the mask to the original image
\[
I_{masked} = I \odot G_O^{(ext)}
\]

Our methodology systematically explores the domain generalisation capabilities of vision transformers through a structured approach. We begin with a feasibility study which demonstrates the resilience of transformer-based architectures against OOD benchmarks. This study reveals BEIT as the most promising model due to its superior performance, attributed to self-supervised learning and masked image modelling.

Building on these findings, we fine-tune the BEIT model on PACS, Office-Home, and 
 DomainNet to ensure a comprehensive evaluation across small, medium, and large-scale datasets. Our fine-tuning process involves feature extraction, attention-based analysis, and domain-specific adaptations while employing best practices like early stopping and model checkpointing. Additionally, we integrate an attention distance analysis to understand BEIT’s information aggregation behaviour by shedding light on local vs. global attention patterns.


\section{Results}\label{sec:Results}

We now present an analysis of the experimental outcomes, and highlight the model’s effectiveness across benchmarks. 

\begin{table*}[ht]
\caption{Experiments with domain generalisation and domain-specific methods} 

\centering 
\begin{tabular}{lcccl}\toprule& \multicolumn{2}{c}{PACS} & \multicolumn{2}{c}{Office-Home}
\\
\midrule
Models           & ~~~Validation~~~  & ~~~Target~~~    & ~~~Validation~~~  & ~~~Target~~~ \\\midrule
GroupDRO    & 0.95 & 0.73 & 0.82 & 0.52   \\
ANDMask & 0.95 & 0.72 &0.81 & 0.44   \\
Mixup & 0.97 & 0.72 & 0.83 & 0.53  \\
MMD   & 0.94 & 0.69 & 0.82 & 0.52  \\
DANN   & 0.94 & 0.73 & 0.83 & 0.51  \\
CORAL   & 0.95 & 0.77 & 0.84 & 0.55  \\
VREx   & 0.97 & 0.80 & 0.76 & 0.49  \\
RSC   & 0.97 & 0.77 & 0.83 & 0.50  \\
ERM   & 0.97 & 0.78 & 0.84 & 0.57  \\
\midrule 
\color{blue}AlexNet   & 0.74 & 0.45 & 0.56 & 0.30  \\
\color{blue}VGGNet16   & 0.80 & 0.47 & 0.50 & 0.23  \\
\color{blue}ResNet18   & 0.86 & 0.51 & 0.65 & 0.52  \\
\color{blue}ResNet50   & 0.89 & 0.57 & 0.70 & 0.62  \\
\color{blue}InceptionV3   & 0.90 & 0.55 & 0.68 & 0.66 \\
\color{blue}DenseNet121   & 0.86 & 0.44 & 0.62 & 0.35 \\
\color{blue}SqueezeNet   & 0.80 & 0.50 & 0.54 & 0.29 \\\bottomrule
\label{tab2} 
\end{tabular}

\end{table*}
Table  \ref{tab2} is taken from~\cite{10.1007/978-3-031-37963-5_60} where we tested the accuracy of two types of models. Results in back font indicate the domain generalised methods and blue  indicates  conventional domain-specific methods. The table has five columns with a main focus on accuracy of validation and unseen target distribution. The initial experiment is conducted on two different benchmarks including PACS and Office-Home, each with four different domains. For a fair comparison, these models have been trained and tested with similar parameters and settings. Moreover, if a model has around 90\%\ accuracy then we can call it a well-trained model. 

The domain generalised methods  overall show better performance for both kinds of benchmarks compared to conventional deep learning models. In both benchmarks, VREx has better performance than others because it has a smaller gap between validation and target distribution. Ideally, a model will be better generalised if the difference between both parameters is as low as possible. Another important insight we can extract from the results in Table~\ref{tab2} is related to a trend of low validation and target accuracy for Office-Home compared to PACS. The main reason for this outcome could be related to the complexity of the benchmark as it has 65 classes with four domains whereas PACS only has 7 classes. 

Furthermore, overall traditional domain-specific methods have less accuracy for both benchmarks.
For PACS, the ResNet50 and InceptionV3 models have  better scores than others for validation and target distributions and they reflect similar behaviours for Office-Home. This  encourages us to use ResNet50 as the baseline for domain generalised methods.  

\subsection{BEIT Results for PACS, Office-Home, and DomainNet Benchmark} \label{Res_Sub1:FT}
\begin{table*}[htbp]
\caption{Results of  fine-tuning experiments using BEIT on three benchmarks 
PACS, Office-Home, DomainNet}
\begin{center}
\begin{tabular}{lcccccc}
\toprule
\textbf{Benchmarks} & \textbf{\textit{Validation Top1 Acc}}& \textbf{\textit{Target Top1 Acc}}& \textbf{\textit{Validation Top5 Acc}}& \textbf{\textit{Target Top5 Acc}}& \textbf{\textit{Gap}}& \textbf{\textit{Precision}} \\
\midrule
PACS & 0.96 & 0.94 & 1.0 & 0.9980 & 0.02 & 0.9464 \\
Office-Home &  0.8597 & 0.8691 & 0.9948 & 0.9679 &-0.0094 & 0.8754\\
DomainNet & 0.7019 & 0.6978 & 0.9347 & 0.8793 & 0.0041 & 0.7111\\
\bottomrule
\multicolumn{5}{l}{}
\end{tabular}
\label{tab3}
\end{center}
\end{table*}

Tables~\ref{tab3}, \ref{tab4}, and \ref{tab5} present  results of fine-tuning the BEIT model in different prospectives. 
Table~\ref{tab3} presents  results discovered after  training and indicates significant improvement in the accuracy metrics. Table~\ref{tab3} indicates top1 and top5 accuracy, gap with IID and OOD, and precision. Models reveal an increase in accuracy for PACS and Office-Home however for DomainNet the model was unable to improve after 70\%\ during the validation testing. In this work, we consider validation distribution as IID and target distribution as OOD which were unseen to models.  

Major reasons behind the better performance for PACS and Office-Home are that BEIT-based fine-tuned vision transformer models could know about similar classes during theie pre-training in the original work. Additionally, PACS and Office-Home are relatively less complex and smaller benchmarks compared to DomainNet. In this study, based on the features of benchmarks~\cite{riaz2023domain} like complexity, number of domain, and sample size, we consider PACS as a basic level benchmark, Office-Home as a medium level benchmark, and DomainNet as a high-level benchmark. As we know that to get better domain generalisation, a model's goal is to decrease the gap between IID and OOD. Therefore, for PACS, we observe a gap of only 2\%\ and for Office-Home  the gap is negative means the model  performs even  better on OOD distribution.

In the case of DomainNet, even though precision is lower than the other two benchmarks, the gap between IID and OOD remains smaller which could be proof of the stability and better generalisation potential of vision transformer-based methods. Table~\ref{tab3} presents the overall results of models on benchmarks yet it does not provide a complete picture.
Therefore, in  Table \ref{tab4} we computed similar metrics for each domain individually.

\begin{table*}[htbp]
\caption{Accuracy and loss for PACS, Office-Home, and DomainNet for each domain independently}
\begin{center}

\begin{tabular}{lccccccc}
\hline 
\textbf{PACS} & \textbf{\textit{Photos}}& \textbf{\textit{Art}}& \textbf{\textit{Cartoon}}& \textbf{\textit{Sketch}}& \textbf{\textit{Metrics}} \\
\hline

\textbf{BEIT} & 0.9766 & 0.9183 & 0.9578 & 0.9371 & accuracy\\
&  0.0493 & 0.2507 & 0.1206 & 0.2227 & loss\\
\hline
\textbf{Office-Home} & \textbf{Art} & \textbf{Clipart} &\textbf{Product}  &\textbf{Real World}  &\textbf{metrics} \\
\hline
\textbf{BEIT} & 0.7979 & 0.8488 & 0.9324 & 0.8645 & accuracy\\
& 0.7443 & 0.5947 & 0.2502 & 0.5339 & loss\\

\hline
\textbf{DomainNet} & \textbf{Clipart} & \textbf{Infograph} & \textbf{Painting} &\textbf{Quickdraw} & \textbf{Real World} & \textbf{Sketch} &\textbf{metrics} \\
\hline
\textbf{BEIT} & 0.7822 & 0.3812 & 0.6893 & 0.6727 & 0.8073 & 0.6764 & accuracy\\
& 0.9129 & 3.0639 & 1.4036 & 1.2023 & 0.7915 & 1.4661 & loss\\
\hline
\multicolumn{5}{l}{}
\end{tabular}
\label{tab4}
\end{center}
\end{table*}

Table \ref{tab4} expresses accuracy and loss for OOD benchmarks according to each domain. The first section of the table highlights performance for BEIT-PACS which has a lower accuracy of 91\%\ and higher loss of 25\%\ for the artwork domain compared to photos, cartoon, and sketch. Moreover, BEIT-PACS show highest accuracy of 97\%\ for the real photos domain which is because the original baseline model of BEIT was trained with similar real images and weights of the pre-trained model could already have better representation for such domains. In the same way, Table~\ref{tab4} also has information about the BEIT-Office-Home model, and the second row of the table focuses on that. Office-Home also has four different domains including art, clipart, product, and real world. 

Interestingly, the BEIT-Office-Home model achieves the highest accuracy of 93\% for the product domain and the real-world domain has the second highest accuracy of 86\%. The loss metrics also indicate the same trend in the table. As a result, we can ask the question of why BEIT-Office-Home could not have the highest accuracy for the real world domain like BEIT-PACS. The answer is that the product domain contains images of various products classes with white backgrounds meaning fewer distractions for models. On the other hand, in the real world domain, images are relatively complex, which means more distraction and closer to reality. As a result,  BEIT-Office-Home has more refined accuracy for the product domain than the real world. Meanwhile, BEIT-Office-Home also exhibits lower accuracy of 79\% and higher loss for the art domain which means it follows the same trends as the BEIT-PACS model.  

Table~\ref{tab4} has information related to DomainNet with six domains including clipart, infograph, painting, quickdraw, real world, and sketch. BEIT-DomainNet highlights similar patterns to the previous two models and our observation stands correct as this model also performs well on real world images with higher accuracy of 80\%\ for the real-world domain because of the same reason we described earlier in both cases before. Conversely, BEIT-DomainNet model performs poorly on the infograph domain as this data distribution is much more diverse and  complex than the original baseline model and even after fine-tuning  it was still extremely challenging to get a better result. 


\begin{table*}[ht]
\caption{Comparison between our trained model and  state-of-the-art methods for OOD generalisation } 
\centering 
\begin{tabular}{lccccccc}\toprule& \multicolumn{3}{c}{PACS} & \multicolumn{3}{c}{Office-Home}& \multicolumn{1}{c}{DomainNet}
\\
\hline
Models           & IID Accuracy   &  OOD Accuracy & Gap   &  IID Accuracy & OOD Accuracy   &  Gap & Target Accuracy \\\hline
GroupDRO& 0.95 & 0.73 & 0.22 & 0.82 & 0.52 & 0.30 & 0.337   \\
ANDMask & 0.95 & 0.72 &0.23 & 0.81 & 0.44 & 0.37 & *  \\
Mixup   & 0.97 & 0.72 & 0.25 & 0.83 & 0.53&0.30&  0.396\\
MMD     & 0.94 & 0.69 & 0.25 & 0.82 & 0.52 &0.30 &  0.394\\
DANN    & 0.94 & 0.73 &0.21 & 0.83 & 0.51 &0.32&  0.384\\
CORAL   & 0.95 & 0.77 &0.18 & 0.84 & 0.55&0.29&  0.418\\
VREx    & 0.97 & 0.80 &0.17 & 0.76 & 0.49&0.27&  0.336\\
RSC     & 0.97 & 0.77 & 0.20 & 0.83 & 0.50 &0.33 & 0.389 \\
ERM     & 0.97 & 0.78 &0.19 & 0.84 & 0.57 &0.27&  0.412\\
\color{blue}Our model   &\color{blue}0.96 & \color{blue}0.94 & \color{blue}0.02 & \color{blue}0.86 &\color{blue} 0.87 &\color{blue} -0.0094 & \color{blue}0.70 \\
\bottomrule
\label{tab5} 
\end{tabular}
\end{table*}

To compare our method with the state-of-the-art, Table~\ref{tab5} presents the performances of various CNN-based domain generalised algorithms from different areas of machine learning including generative adversarial learning, augmentation, invariant feature learning, meta-learning, and lifelong learning. 
We present experiments for PACS and Office-Home and in case of DomainNet we present performance figures taken from~\cite{gulrajani2021in}. The columns in the table are the model names, IID accuracy, OOD accuracy, and gap for PACS and for Office-Home. For DomainNet, our work here only considers target accuracy. It is clear from the table that our method outperforms the state-of-the-art approaches in all benchmarks with a substantial difference. In the case of PACS and Office-Home, these CNN backboned algorithms have few things in common. For example, their in-domain accuracy is comparatively high which could be better for tasks where models have to tackle  partial domain shifting. Nevertheless, the performance of these methods on OOD accuracy is poor and as a result, such methods have larger gaps in their performances. To be specific, other algorithms have on average 21.1\%\ and 30.5\%\ gaps for PACS and Office-Home respectively.   

Our method accomplishes its task by reducing the gap and also maintaining  IID and OOD accuracy. If we examine the performance figures for PACS, our method obtains an overall 96\%\ and 94\%\ for IID and OOD accuracy respectively and the gap shrinks from 21.1\%\ to 2\%\. ~Our method also shows similar trends for Office-Home and reduces the gap parameter. The gap is negative because our model perform slightly better on OOD than on IID which is also a good sign for domain generalisation capabilities. Table~\ref{tab5} also shows that overall the target accuracy is not that high for all approaches but still our method outperforms all existing approaches with a difference of 37.98\%\ if we consider 31.8\%\ as average accuracy.  


\subsection{Results for Newly Generated OOD Benchmarks}
To further measure the domain generalisation potential of our model, Tables~\ref{paper6_tab1}, \ref{paper6_tab2}, and \ref{paper6_tab3} indicate the results on newly generated OOD benchmarks using GroundingDino and SAM methods on three benchmark datasets. Using these  model architectures, we calculated the approximate segment mask of objects present in  images and created a periodic grid masking in the area of shape as shown in Figure~\ref{paper6_fig3}. To test the OOD potential of our model, the method created  data distributions with different occlusion ratios of 25\%, 50\%, and 75\% masking of objects as seen in  Figure~\ref{paper6_fig3}. We performed these transformations on all three  benchmarks and created OOD distributions that models have not used in their training.

\begin{table*}[!htb]
\caption{Results on newly generated OOD PACS benchmark using GroundingDino and SAM for image augmentation.}
\begin{center}
\renewcommand{\arraystretch}{1.3}
\begin{tabular}{lccccc}
\toprule
\textbf{Number of Grids}&\textbf{Dataset} & \textbf{\textit{Top1 Acc}}& \textbf{\textit{Top5 Acc}}& \textbf{\textit{Top1 Gap}}& \textbf{\textit{Top5 Gap}} \\
\midrule
&PACS-original & 0.9400 & 0.9980 & -& - \\
\midrule
\multirow{3}{*}{2 Grids} &
PACS-occluded$_{25\%}$ &  0.9008 & 0.9921 & -0.04 & -0.01 \\
&PACS-occluded$_{50\%}$ & 0.7804 & 0.9822 & -0.16 & -0.02\\
&PACS-occluded$_{75\%}$ & 0.7156 & 0.9744 & -0.22 & -0.02  \\
\midrule 
\multirow{3}{*}{5 Grids} &
PACS-occluded$_{25\%}$ & 0.8307 & 0.9813 & -0.11 & -0.02 \\
&PACS-occluded$_{50\%}$ & 0.6616 & 0.9379 & -0.28 & -0.06\\
&PACS-occluded$_{75\%}$ & 0.6589 & 0.9517 & -0.29 & -0.05  \\
\midrule 
\multirow{3}{*}{9 Grids} &
PACS-occluded$_{25\%}$ & 0.6295 & 0.9576 & -0.31 & -0.04 \\
&PACS-occluded$_{50\%}$ & 0.4103 & 0.9546 & -0.53 & -0.04\\
&PACS-occluded$_{75\%}$ & 0.5182 & 0.9428 & -0.42 & -0.06  \\
\midrule 
\multirow{3}{*}{5 Grids outside segments edges} &
PACS-occluded$_{25\%}$ & 0.5699 & 0.9379 & -0.3701 & -0.0601 \\
&PACS-occluded$_{50\%}$ & 0.2800 & 0.7959 & -0.6600 & -0.2021\\
&PACS-occluded$_{75\%}$ & 0.2527 & 0.7800 & -0.6873 &  -0.218 \\
\bottomrule
\multicolumn{6}{l}{}
\end{tabular}
\label{paper6_tab1}
\end{center}
\end{table*}

Table~\ref{paper6_tab1}, like the other results tables in this paper, has six columns indicating the number of grids, the benchmark dataset, the top 1 and top 5 accuracy, and the gaps in accuracy between the PACS benchmark with occlusion, and PACS-original. The top 1 gap is calculated with the difference between the original top 1 score and occluded data distribution scores while the top 5 gap is calculated between the original top 5 accuracy metrics of the testing benchmark and the top 5 accuracy of newly generated benchmarks.  Table~\ref{paper6_tab1} demonstrates that when we occluded 25\% of shape which is shown in the first row of Figure~\ref{paper6_fig3}, our PACS-BEIT model still performs well on the PACS-occluded$_{25\%}$ OOD benchmark. The top 1 gap only increased by 8\% from the original score and the top 5 gap is at 2\%. This is a significant achievement for the model to show its resilience against outside noise/distractions. For PACS-occluded$_{50\%}$ and PACS-occluded$_{75\%}$, examples can be seen from Figure~\ref{paper6_fig3} that the transformer-based PACS-BEIT model decreases its accuracy to 28\% and 29\% 
from the original benchmark, respectively. Furthermore,it is interesting to note that even if 75\% of the shapes are blocked by grid masks the model is stabilised at around 65\% accuracy. Even though the top 1 accuracy has a declining trend for 50\% and 75\% occluded images in the dataset, the top 5 accuracy is still close to the original top 5 scores.

Table~\ref{paper6_tab1}  shows the transition of accuracies for the PACS benchmark when we increase the number of grids. Ultimately, this means reducing the grid size, and Figure~\ref{paper6_fig3} shows what each occluded ratio looks like in each case. The main reason we conducted experiments based on a number of grids is that we would like to identify the overall trend of the model and what would be the optimal grid size to most impact the performance of the model and in which scenario the generated PACS benchmark provides stable results regardless of masking of visual information. For 2 grids the PACS-BEIT model shows the most significant results compared to other implemented scenarios. Even though we have occluded 75\% of shapes we still get 72\% accuracy and only a 22\% decrease compared to the original scores. When we block 25\% of shapes a 4\% decrease is observed in the top 1 score. 

We then increased the number of grids to 5 which means reducing the overall size of the grid unit model slightly dropping performance compared to the 2-grid benchmark. For 25\%, 50\%, and 75\% occlusion ratios it drops 11\%, 28\% and 29\% accuracy, respectively. To understand the reasons behind this  we further  decrease the sizes of the grids and the findings are unique to observe. For example, when we block smaller chunks of information in  images, the PACS-BEIT model seems to suffer and significantly drop the accuracy parameter which means domain generalisation is most affected when we block smaller chunks of information with respect to bigger chunks. Therefore, for 9 grids, a 31\% drop is observed for the 25\% occluded benchmark images. Interestingly,  another trend  appears namely that when we increase the number of grids, PACS-occluded$_{50\%}$ performs poorer than PACS-occluded$_{75\%}$.

\begin{table*}[!htb]
\caption{Results on newly generated OOD benchmarks using zero-shot instance masking for Office-Home.}
\begin{center}
\renewcommand{\arraystretch}{1.3}
\begin{tabular}{lccccc}
\toprule
\textbf{Number of Grids}&\textbf{Dataset} & \textbf{\textit{Top1 Acc}}& \textbf{\textit{Top5 Acc}}& \textbf{\textit{Top1 Gap}}& \textbf{\textit{Top5 Gap}} \\
\midrule
&Office-Home-original & 0.8691 & 0.9679 & - & - \\
\midrule
\multirow{3}{*}{2 Grids} &
Office-Home-occluded$_{25\%}$&  0.8203& 0.9442 & -0.05 & -0.02 \\
&Office-Home-occluded$_{50\%}$ &  0.7244  & 0.9034 & -0.14 & -0.06 \\
&Office-Home-occluded$_{75\%}$ &  0.6692 & 0.8556 & -0.20 & -0.11 \\
\midrule 
\multirow{3}{*}{5 Grids} &
Office-Home-occluded$_{25\%}$&  0.7987 & 0.9355 & -0.07 & -0.03 \\
&Office-Home-occluded$_{50\%}$ &  0.6007 & 0.7956 & -0.27 & -0.17 \\
&Office-Home-occluded$_{75\%}$ &  0.5810 & 0.7777 & -0.29 & -0.19 \\
\midrule 
\multirow{3}{*}{9 Grids} &
Office-Home-occluded$_{25\%}$&  0.6297 & 0.8347 & -0.18 & -0.13 \\
&Office-Home-occluded$_{50\%}$ &  0.5655 & 0.7796 & -0.30 & -0.19 \\
&Office-Home-occluded$_{75\%}$ &  0.5338 & 0.7530 & -0.33 & -0.21 \\
\midrule 
\multirow{3}{*}{5 Grids outside segments edges} &
Office-Home-occluded$_{25\%}$ & 0.5484 & 0.7844 & -0.3207 & -0.1835 \\
&Office-Home-occluded$_{50\%}$ & 0.2523 &0.4472  & -0.6168 &-0.5207 \\
&Office-Home-occluded$_{75\%}$ & 0.1799 & 0.3820 & -0.6892 &  -0.5859 \\
\bottomrule
\multicolumn{6}{l}{}
\end{tabular}
\label{paper6_tab2}
\end{center}
\end{table*}

Table \ref{paper6_tab2} offers similar results for the Office-Home benchmark. In the case of 2 grids, Office-Home is also less affected by the masking of shapes. It also provides proof of concept that when we corrupt big chunks of information, at the same time we also allow the model to see other big chunks to make  final predictions and remain less affected and utilise the available information in the best way. For instance, with 2 grids, the Office-BEIT model also shows similar patterns to  PACS-BEIT and shows 5\%, 14\% and 20\% drops in performance compared to the original baseline results.

For 5 grids, the Office-Home-occluded$_{25\%}$ has a 7\% top 1 gap meaning the model decreases only 7\% in accuracy compared to the original baseline results and only 3\% for the top 5 gap score. In the case of masking 50\% and 75\%, the Office-Home-BEIT model remains at 60\% and 58\% top 1 accuracy results. This is  surprising because even though Office-Home is a medium-level benchmark compared to PACS, and the model performs well, it explains the potential of the model to have better generalisation. The top 1 gap is increased from 7\% to 27\% for the 50\% occluded benchmark and even if we increased the occlusion rate by 25\% to  Office-Home-occluded$_{75\%}$, the model only shows a 29\% drop which is 2\% more than the Office-Home-occluded$_{50\%}$. A similar trend can be seen in the top 5 gap scores. The overall behaviour conveys an important message about learning generalised connections for the classes present in the images and being less affected by domain-shifting mechanisms. 

Similar to the benchmarking results from PACS, when we enhance the number of grids to 9, a similar declining trend can be observed from Table~\ref{paper6_tab2}. To the best of our understanding, this means that when we block bigger chunks of input images, the model is still able to use the relationship between different parts of an image at feature levels and when we increase the number of grids, it becomes harder for the model to maintain its performance. 

\begin{table*}[!htb]
\caption{Results on newly generated OOD benchmarks using zero shot instance masking using SAM and Grounding DINO for DomainNet}
\begin{center}
\renewcommand{\arraystretch}{1.3}
\begin{tabular}{lccccc}
\toprule
\textbf{Number of Grids}&\textbf{Dataset} & \textbf{\textit{Top1 Acc}}& \textbf{\textit{Top5 Acc}}& \textbf{\textit{Top1 Gap}}& \textbf{\textit{Top5 Gap}} \\
\midrule
& DomainNet-original & 0.6978 &  0.8793 & - & - \\
\midrule
\multirow{3}{*}{2 Grids} &
DomainNet-occluded$_{25\%}$ & 0.6197 & 0.8283 & -0.0781 & -0.051\\
&DomainNet-occluded$_{50\%}$ &0.5226  & 0.7407 & -0.1752 & -0.1386\\
&DomainNet-occluded$_{75\%}$ &0.4805  & 0.6916 & -0.2173 & -0.1877 \\
\midrule 
\multirow{3}{*}{5 Grids} &
DomainNet-occluded$_{25\%}$ & 0.5898 & 0.8050 & -0.10 & -0.07\\
&DomainNet-occluded$_{50\%}$ & 0.4322 & 0.6361 & -0.26 & -0.24\\
&DomainNet-occluded$_{75\%}$ & 0.4326 & 0.6365 & -0.26 & -0.24 \\
\midrule 
\multirow{3}{*}{9 Grids} &
DomainNet-occluded$_{25\%}$ & 0.4882 & 0.7031 & -0.21 & -0.18\\
&DomainNet-occluded$_{50\%}$ & 0.3943 & 0.5840 & -0.30 & -0.3\\
&DomainNet-occluded$_{75\%}$ & 0.4094 & 0.6038 & -0.29 & -0.28 \\
\midrule 
\multirow{3}{*}{5 Grids outside segments edges} &
DomainNet-occluded$_{25\%}$ & 0.3527 & 0.5759 & -0.3451 & -0.3034 \\
&DomainNet-occluded$_{50\%}$ & 0.1596 & 0.2714 & -0.5382 & -0.6079\\
&DomainNet-occluded$_{75\%}$ & 0.1542 & 0.2533 & -0.5436 & -0.6260  \\
\bottomrule
\multicolumn{6}{l}{}
\end{tabular}
\label{paper6_tab3}
\end{center}
\end{table*}

The results for DomainNet-occluded$_{25\%}$ shown in Table~\ref{paper6_tab3} show an approximately 10\% reduction in top 1 accuracy compared to baseline scores and when we increase the occlusion rate to the 50\% and 75\%, the  model shows 26\% overall reduction in top 1 accuracy. The top 5 gap remains at 24\% for both cases. As we know, DomainNet is one of the more complex and larger domain generalisation benchmarks but in terms of generalisation concerns, the performance of the trained model did not drop regardless of domain shifting.

At the last rows of each table, it can observed that for such data distributions with shapes outside the shapes or edges, all three models show decreases in their performance. This also highlights another  point which is that our learned models focus on shapes as we stated earlier. Therefore, when our methods create distributions with grids outside the segmented edges this deforms the shapes of objects and this has the greatest  effect on the generalisation of models.

\section{Analysis of Attention and Discussion }
\label{sec:Attention}
We now present an analysis of domain generalisation with the BEIT type vision transformer for  selected datasets. We consider self-attention distance metrics to explore different data distributions, using the learned latent space of each model for individual domains, and we investigate which domain is not effectively learned by the models. We also illustrate some sample attention maps with token masks and their respective attention heat maps.

\subsection{Self-Attention Distance Analysis for Domain Generalisation}
\begin{figure*}[htbp]
\centering
\includegraphics[width=0.8\textwidth]{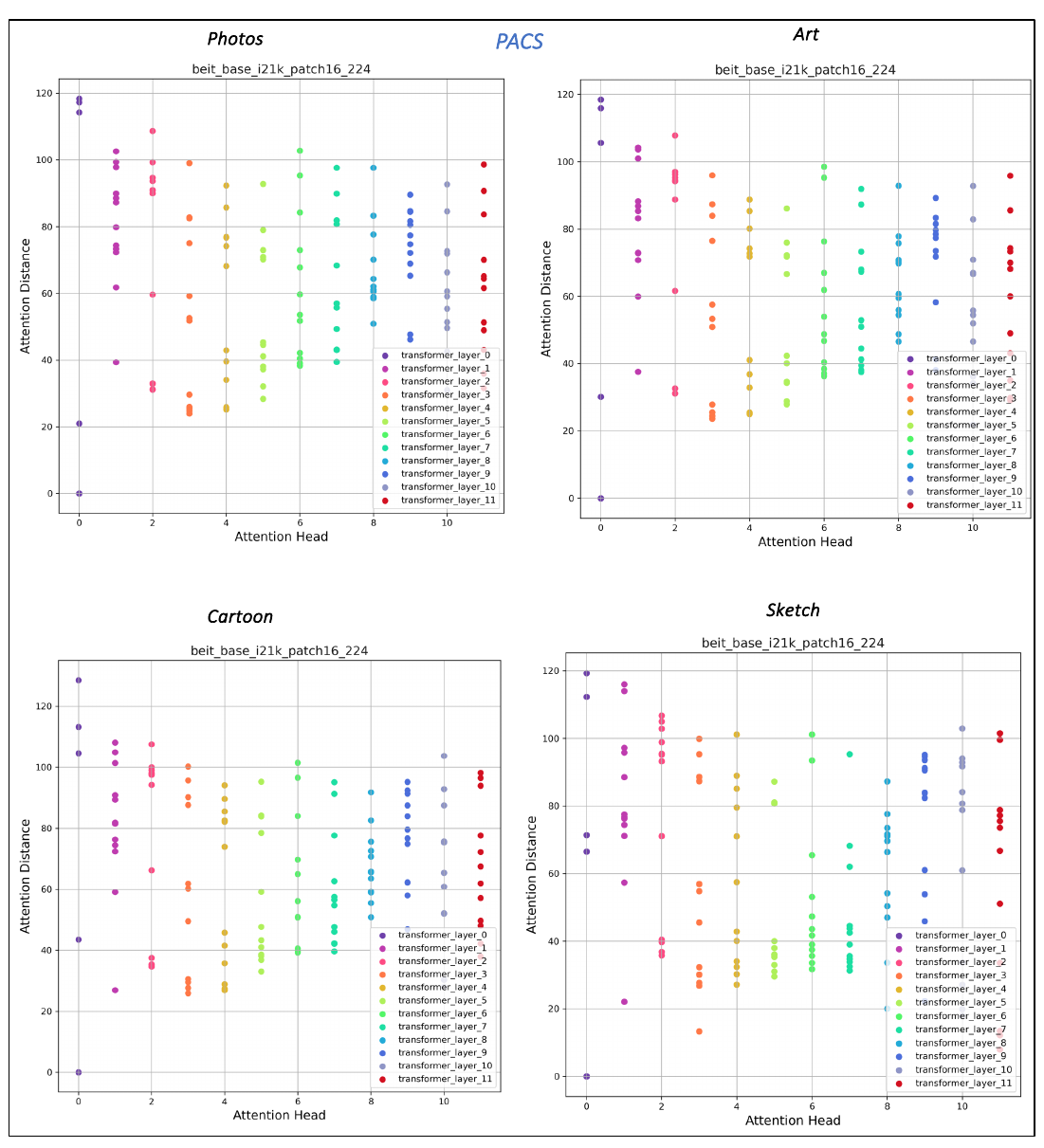} %
\caption{Mean self-attention distance analysis computed by our model when tested on the PACS benchmark.}
\label{paper5_fig4}
\end{figure*}
After fine-tuning a model it is interesting to calculate the mean attention distance of the learned latent space because this provides insights about layers of the model such as which layer learns most about local and global information. 
%
%
Our interest here comes from the literature which showed that if a model has a mixture of local and global spatial information in its early layers, this is a positive indication for the model to perform well for domain generalisation~\cite{NEURIPS2021_652cf383}. In addition, ideally in later layers a vision transformer model should only focus on global information which points to  the insights shown in~\cite{wang2019learning}.
In this way, the learned features will be a combination of very strong and diversified local and global features.   
Similar to \cite{NEURIPS2021_652cf383}, our method follows steps to calculate the mean attention distance. This is defined as the distance between a query pixel and the rest of the patch, multiplied by attention weights. We compute an attention distance for each head and then average these over all OOD testing benchmarks.  Figures~\ref{paper5_fig4} and \ref{paper5_fig5} show the mean self-attention head distance analysis for PACS and for Office-Home, respectively. We calculate distances for each domain separately. 

\subsection{Self Attention Distance Analysis for the Latent Space of PACS, Office-Home and DomainNet}

\begin{figure*}[htbp]
\centering
\includegraphics[width=0.8\textwidth]{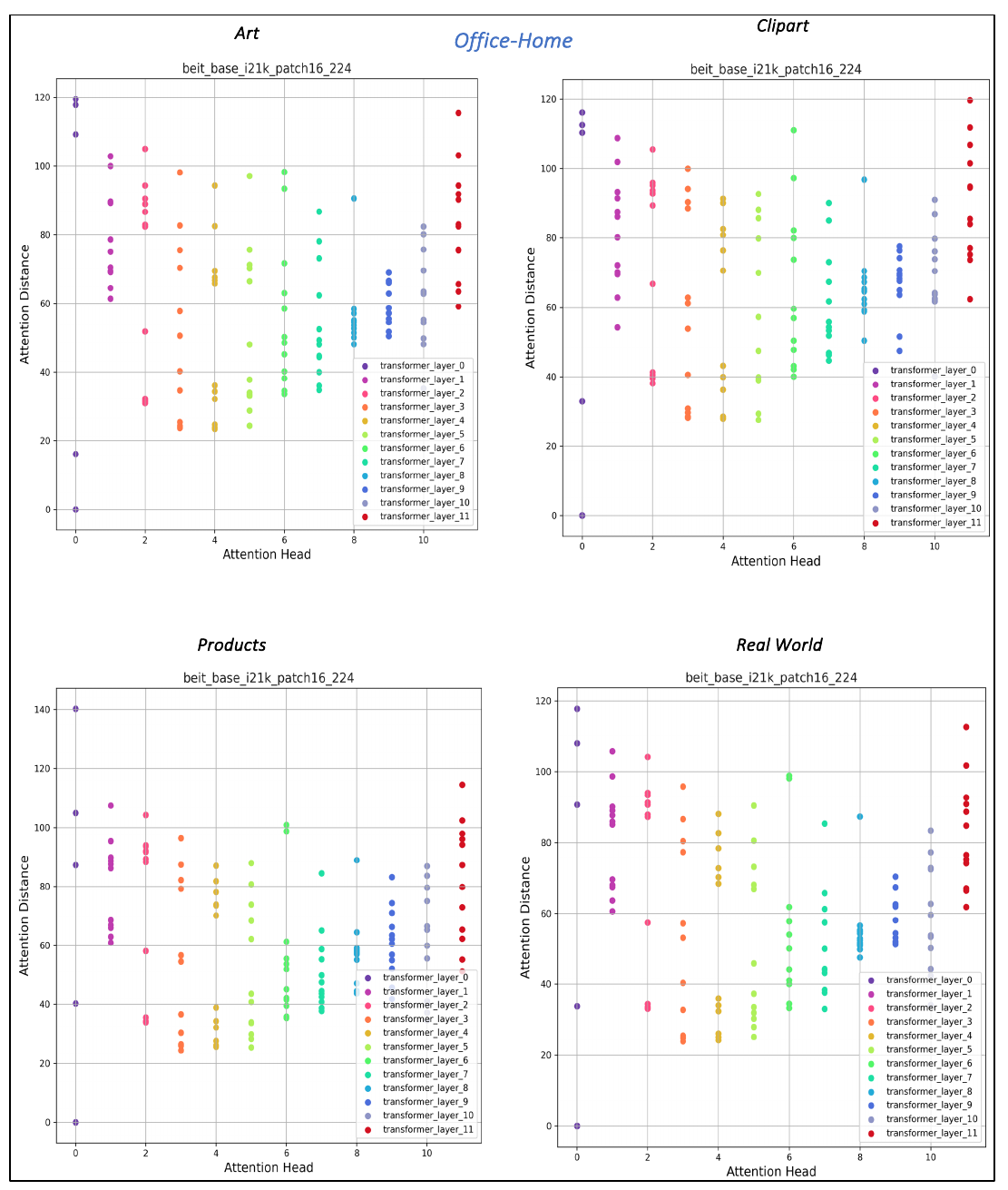} %
\caption{Mean self-attention distance analysis computed on the Office-Home benchmark.}
\label{paper5_fig5}
\end{figure*}

Figure~\ref{paper5_fig4} presents  four graphs  where each graph represents each domain in PACS. Similarly, Figure \ref{paper5_fig5} represent four graphs for Office-Home according to its  domains. The x-axis of each graph is the attention head and the y-axis is attention distance. Overall, the model shows a similar trend  as the first layer model which has a combination of both local and global distances. Higher distances relate to global information and lower distances correspond to local information. In both benchmarks, for each domain our models start fine-tuning from a mixture of both types of attention and then try to shift the learning focus on to the global side during the later layers. For this reason we can correlate the results of the experiment from Table~\ref{tab5} with this understanding of the graphs. As  the final layers of the  model mainly access higher global distances and this means stable and better generalisation ability. The graphs highlight a possible reason behind the state-of-the-art performance of our fine-tuned model.

\begin{figure*}[htbp]
\centering
\includegraphics[width=0.8\textwidth]{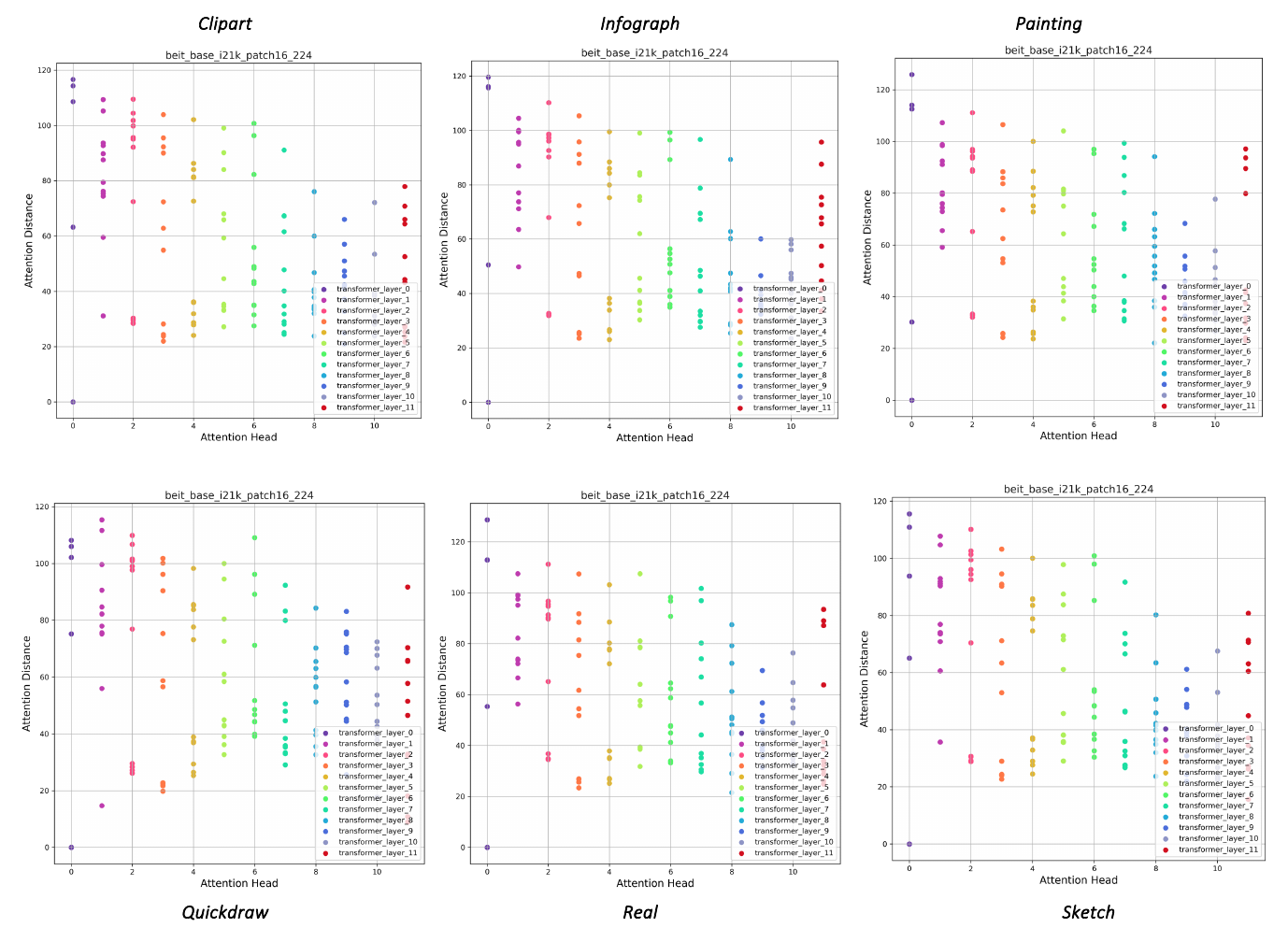} %
\caption{Latent space of self-attention distance for our model learned when tested on DomainNet}
\label{paper5_fig6}
\end{figure*}

In the same manner, Figure~\ref{paper5_fig6} indicates the mean self-attention distances for DomainNet. Overall, the model shows a downward trend for all domains of the DomainNet dataset which means that the model starts fine-tuning the same as PACS and Office-Home but in the later layers the model is unable to properly shift towards the global side of the latent space which we can see by low mean distances in the graphs. Consequently, rather than the model paying attention to global distances it struggles to learn global features and  focuses on  local parts. However, from  Table~\ref{tab5} it is clear that our model still has better accuracy than other available methods but the generalisation ability of our fine-tuned model is not that effective in PACS or Office-Home. 

One of the reasons our models outperforms others  is because the early  and middle layers have a mixture of both the local and global feature space.  
During our analysis of DomainNet, we also observe that after the first few  layers, mean attention distance starts decreasing which is a common trend across all domains. This means we actually do not need to train all layers of the transformers because the model is not learning generalised features and thus we could reduce the number of training layers from the transformers. 

In summary, the attention distance analysis conveys extremely important insights. For instance, if attention distances are low it means that the model is too  focused on local features/information which could lead towards poor domain generalisation. On the contrary, high attention distances mean more global features or information is learned by the model which usually leads models towards better and more robust DG. Figures\ref{paper5_fig4}, \ref{paper5_fig5}, and \ref{paper5_fig6} also deliver similar messages across each domain. 

\begin{figure*}[!ht]
\includegraphics[width=0.9\textwidth]{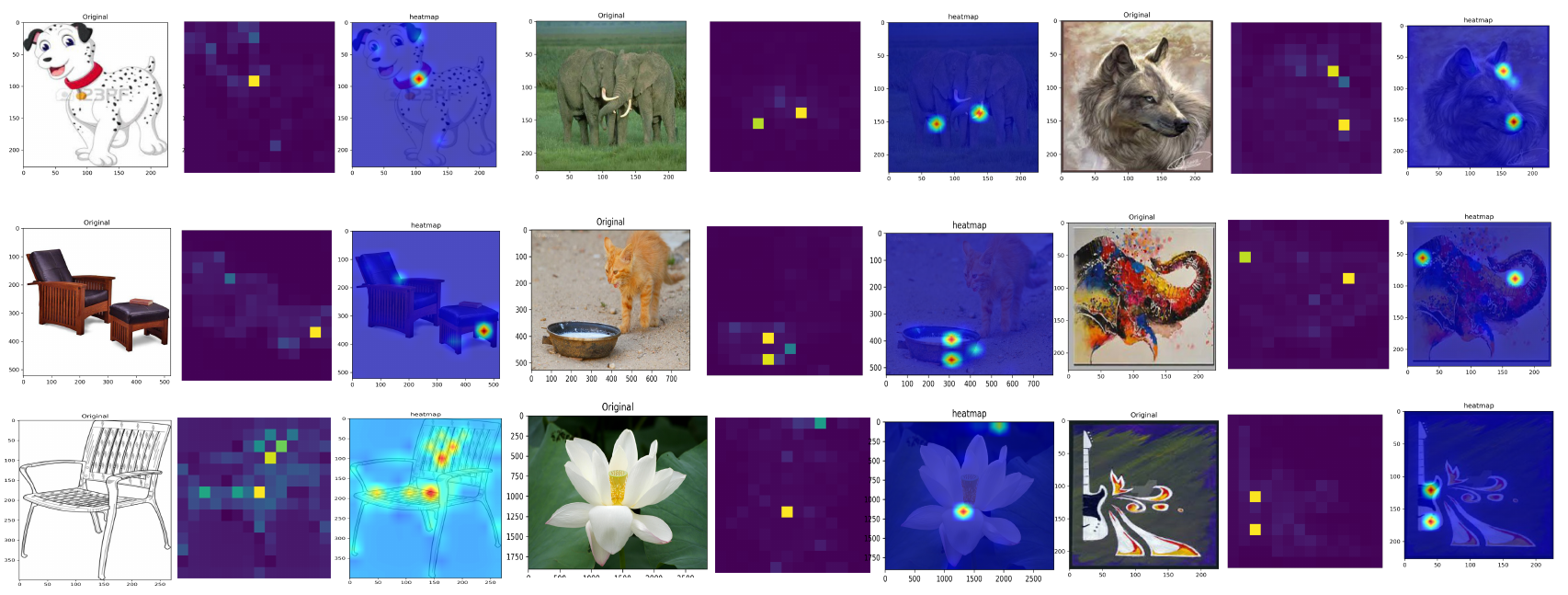} %
\caption{Attention map analysis for for sample images for various domains.}
\label{paper5_fig7}
\end{figure*}

Finbally, it is important to visualise the attention maps learned by models for various domains. In Figure~\ref{paper5_fig7}, we present a number of  original images  (12) with their  token mask extracted directly from the latent space of the fine-tuned model and heatmaps of attention on the original images. It can be  seen from this that model is paying attention to the shapes rather than backgrounds, and is ignoring noise. 

\section{Conclusion and Future Research Directions }
\label{sec:Conclusion}

In pursuit of robust AI systems capable of generalising across unpredictable real-world conditions, this research establishes vision transformers, particularly BEIT, as a paradigm shift in domain generalisation. By integrating self-supervised learning, masked image modelling (MIM), and a global self-attention mechanism, BEIT achieves state-of-the-art performance on benchmarks including PACS (94\% accuracy), Office-Home (87\%), and DomainNet. It reduces the critical gap between in-distribution (IID) and out-of-distribution (OOD) accuracy from 21.1\% to 2\% which is evidence of its ability to prioritise object shapes over textures and learn invariant features.

To  validate these claims we developed a novel framework for generating synthetic OOD benchmarks using zero-shot segmentation (via Segment Anything Modeland Grounding DINO) and controlled grid masking. This approach revealed BEIT’s remarkable resilience: it retained strong performance even with 75\% occlusion of object regions, outperforming CNNs and other vision transformers by up to 37\%. However, when occlusions disrupted object shapes (e.g., grids outside boundaries), performance plummeted by 68\% indicating the importance of structural integrity in DG. These findings empirically validate BEIT’s denoising capabilities and its reliance on global contexts when exposed to vulnerabilities tied to local texture bias in conventional models.

The  contributions of this paper can be summarised as follows:

\begin{itemize}
    \item \textbf{Architectural Insights:} BEIT’s self-attention mechanism and MIM pre-training enable robust feature learning that reduces the reliance on transfer learning for unseen distributions;

    \item \textbf{Synthetic Benchmarking:} A scalable method to stress-test models under different OOD scenarios and evaluation of our pre-trained models;

    \item \textbf{Practical Guidelines:} Prioritise vision transformers with global attention for mission-critical applications (e.g., autonomous systems), and avoid over-reliance on local textures.
\end{itemize}

This work also advances methodological rigour in DG research. By quantifying performance through metrics like IID-OOD gaps and precision, and analysing self-attention distances, we provide a design for evaluating model robustness. The sharp contrast in DomainNet’s performance, where limited global attention in later layers hindered generalisation and highlights the need for hybrid architectures combining BEIT’s strengths with explicit shape-awareness.

For  future research directions, it is important to develop pruning strategies to enhance efficiency without compromising DG. There is also scope for an exploration of hybrid models integrating global reasoning of vision transformers with CNN-like local feature extraction and an extension of synthetic benchmarks is needed which can represents dynamics of real-world environments (e.g., weather changes, motion blur).

In summary, this research demonstrated that how models learn (global vs. local focus) is as critical as what they learn. By leveraging vision transformer's inherent strengths and addressing their vulnerabilities, we move closer to AI systems that are generalised reliably in an ever-changing world, a long-awaited milestone for the machine learning researcher community to deploy trustworthy machine learning in practice.

\section*{Acknowledgments}
This work is funded under the ML-Labs SFI Centre for Researcher Training in Machine Learning (18/CRT/ 6183) and part-funded by SFI (12/RC/2289\_P2) at Insight the Research Ireland Centre for Data Analytics at DCU.

 \bibliographystyle{elsarticle-num} 
 \bibliography{cas-refs}





\end{document}